\documentclass{article}

\PassOptionsToPackage{numbers}{natbib}
\usepackage[preprint]{neurips_2026}
\usepackage{amsmath,amssymb}
\usepackage{cancel}
\usepackage{relsize}
\usepackage{array}
\usepackage{svg}
\newcolumntype{C}[1]{>{\centering\arraybackslash}p{#1}}
\usepackage{makecell}
\usepackage{stfloats}
\usepackage{natbib}
\newcommand{\1}{\mathbbm{1}}
\newcommand{\EE}{\mathbb{E}}

\usepackage[utf8]{inputenc} 
\usepackage[T1]{fontenc}    
\usepackage{hyperref}       
\usepackage{url}            
\usepackage{booktabs}       
\usepackage{amsfonts}       
\usepackage{nicefrac}       
\usepackage{microtype}      
\usepackage{xcolor}         
\newcommand{\Var}{\mathrm{Var}}
\usepackage{bbm}
\usepackage{nomencl}
\usepackage[dvipsnames]{xcolor}
\usepackage{multirow}
\usepackage{placeins}
\usepackage{enumitem}
\usepackage[mathscr]{eucal}
\definecolor{reach}{HTML}{00d192}
\makenomenclature
\usepackage{booktabs}
\usepackage{siunitx}

\usepackage[switch]{lineno}
\usepackage{subcaption}
\usepackage{graphicx}
\usepackage{cleveref}
\title{Efficient Generative Prediction for EHR Foundation Models: The SCOPE and REACH Estimators}

\author{Luke Solo}
\author{%
  Luke Solo \\
  University of Chicago\\
  \texttt{lsolo@uchicago.edu} \\
  \And
  Matthew B. A. McDermott \\
  Columbia University \\
  \texttt{mm6677@cumc.columbia.edu}\\
  \And
  William F. Parker\\
  University of Chicago \\
  \texttt{wparker@uchicago.edu}\\
  \And
  Bashar Ramadan \\
  University of Chicago \\
  \texttt{basharramadan@uchicago.edu}\\
  \And
  Michael C. Burkhart\\
  University of Chicago \\
  \texttt{burkh4rt@uchicago.edu}\\
  \And
  Brett K. Beaulieu-Jones \\
  University of Chicago \\
  \texttt{beaulieujones@uchicago.edu}\\
}

\begin{document}
\maketitle

\begin{abstract}
Generative foundation models trained on tokenized electronic health record (EHR) timelines show promise for clinical outcome prediction via Monte Carlo sampling of simulated future trajectories. However, this approach suffers from three coupled limitations: sparse estimate distributions that poorly differentiate patient risk levels, extreme computational cost, and high sampling variance. We propose two new estimators that leverage next-token probability distributions underutilized by standard Monte Carlo: the Sum of Conditional Outcome Probability Estimator (SCOPE) and Risk Estimation from Anticipated Conditional Hazards (REACH). We prove both are unbiased, that REACH guarantees variance reduction over Monte Carlo for any model and outcome, and that REACH is a Rao-Blackwellization of any naive importance sampling scheme that preserves the non-outcome token distribution. 
Empirically, across $11$ clinically important outcomes in MIMIC-IV and the UChicago health system, SCOPE and REACH match $100$-sample Monte Carlo accuracy with median token reductions of $2.5\times$ to $3.4\times$ and reductions exceeding $80\times$ for the rarest outcomes, with calibration preserved throughout. Because SCOPE reuses a single sampled pool across an arbitrary number of outcomes at no marginal generation cost while REACH provides a per-task variance guarantee, the two estimators are complementary in deployment and together meaningfully reduce the inference budget required for generative EHR foundation models, particularly for rare, high-impact outcomes in healthcare.
\end{abstract}

\section{Introduction}

\begin{figure}[tb]
  \centering
  \includegraphics[width=\linewidth]{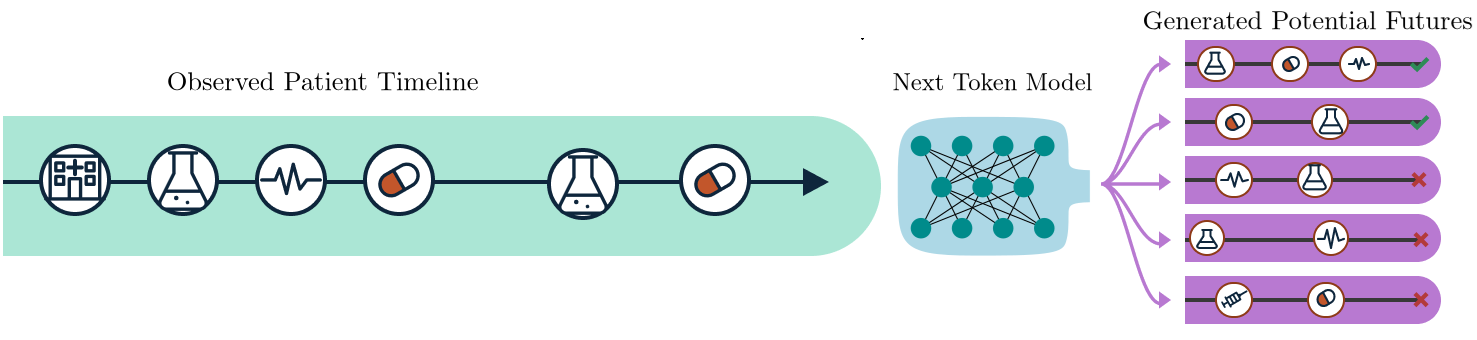}
  \caption{Visualization of the generative outcome prediction framework. Observed tokenized patient timelines are passed to a trained next token prediciton model, which can then be used to generate multiple possible future completions of the initial observed timeline.}
  \label{fig:gen}
\end{figure}

Foundation models pre-trained on sequences of tokenized electronic health record (EHR) data have shown performance and versatility across a range of downstream predictive tasks~\citep{Li20,Ras21,wornow_shaky_2023,Wor23,Kra24,Wax25}. In this framework, the patient's interactions with healthcare are converted from their operational representation (tabular EHR databases) into a sequence of tokens that describe medical events in the patient’s timeline, such as labs, vitals, and prescriptions~\citep{McD23,Wor25,PSB26}. After being trained on the self-supervised task of predicting the next token in a patient’s timeline, these models can auto-regressively generate multiple potential future timelines for a patient. This pool of simulated timelines can then be used to predict an arbitrary number of downstream outcomes by using the proportion of simulated timelines in which a patient experiences the outcome of interest as a risk estimate~\citep[][\figurename{}~\ref{fig:gen}]{Kra24,renc_zero_2024,Ren25,Wax25,Bli24,Aka25,Pan25,Raj25,Fal25}.

Inference by simulating future timelines is often much more computationally demanding than traditional supervised prediction methods, limiting the adoption of the technique. In additional experiments reproducing the pipeline of an existing paper (Appendices [\ref{app:abt_ethos}],[\ref{app:ea_empirical}],[\ref{sec.inf_time_red}],[\ref{app:mortality_perf}], and [\ref{app:icu_adm_perf}]), performing inference to generate future timelines required more than $1500$ A100 GPU hours~\citep{Ren25}. This could be a major barrier to prospective clinical deployment, particularly in acute settings such as the intensive care unit (ICU) where risk estimates may need to be refreshed frequently and with low latency. Without reducing inference-time requirements, the method may be limited to offline or low-frequency use rather than real-time decision support.  

Due to this computational expense, current implementations are limited in the number of future timelines they can sample per patient. This is particularly problematic because the granularity of this estimator is directly proportional to the number of sampled timelines. For rare outcomes, this will place the true risk of patients with very different risk scores between the same two lowest possible risk scores. For example, if an important outcome has a prevalence of 1/10,000, a Monte Carlo estimator with even $100$ sampled futures will fail to assign a higher risk score to a patient with a risk $10$ times higher than an average patient over $90\%$ of the time (see Appendix~\ref{app:dispersion}).  This arithmetic limitation is misaligned with clinical needs: the most consequential decisions are often driven by low-prevalence but high-impact outcomes, where even small absolute differences in risk can meaningfully change diagnostic and therapeutic actions (e.g. evaluating chest or back pain for aortic dissection \citep{hsia_national_2016}). 

In response to these challenges, we propose two ways of converting generated patient futures into outcome probability estimates that decouple the granularity of the estimator from the number of sampled timelines, allowing for accurate prediction of rare outcomes using much smaller computational budgets: Sum of Conditional Outcome Probability Estimates (\textbf{SCOPE}) and Risk Estimation from Anticipated Conditional Hazards (\textbf{REACH}). In particular, SCOPE can be computed for any number of outcomes at exactly the same token generation cost as traditional Monte Carlo estimation. REACH, on the other hand, requires a small task specific generation cost proportional to the prevalence of the outcome of interest. The task-marginal generation cost of REACH is therefore lowest for the rarest outcomes, where these methods provide the greatest increases in predictive accuracy. REACH is also demonstrated formally to reduce estimator variance over both the Monte Carlo and SCOPE methods, guaranteeing a more precise estimator given an equal number of samples. 

We summarize our contributions as follows:
\begin{itemize}
    \item \textbf{Continuous risk scoring} \textit{(SCOPE \& REACH).} We leverage next-token probability distributions to produce granular, continuous risk scores per timeline, resolving estimator dispersion and enabling accurate rare-outcome prediction with substantially fewer sampled futures.
    \item \textbf{Theoretical grounding} \textit{(SCOPE \& REACH).} We provide formal derivations, proofs of unbiasedness, and closed-form characterization of estimator variance.
    \item \textbf{Fixed-cost multi-outcome estimation} \textit{(SCOPE).} A single pool of sampled futures suffices to compute SCOPE estimates for an arbitrary number of outcomes simultaneously, with no additional inference cost beyond what is required for a standard MC estimate.
    \item \textbf{Guaranteed variance reduction} \textit{(REACH).} REACH provably reduces variance relative to both MC and SCOPE across all model-task combinations, at a marginal inference cost proportional to outcome prevalence, with rarer outcomes incurring the least cost.
\end{itemize}

\subsection{Problem Statement}
In a standard workflow, electronic health records are represented as token sequences. Each sequence begins with demographic tokens, followed by chronologically ordered tokens for medical events such as: vitals, labs, medications, and procedures. A generative model $P$ is then trained on these sequences using using a standard language modeling objective (next-token prediction). 

At inference time, given a partial sequence, the model conditions on all observed tokens up to a prediction time and autoregressively samples $n\ge 1$ possible continuation sequences $X^{(1)}, X^{(2)}, \dotsc, X^{(n)}$.

Let $O$ denote a token corresponding to an event of interest. For a sequence $X$, define $T_O(X)$ as the first index at which $O$ appears, or $\infty$ if it does not occur. Similarly, let $T_E(X)$ denote the first index at which the outcome window has been exceeded (for example, the position of a discharge token). We assume sequences are finite ($T_E(X)<\infty$) and that $T_E \ne T_O$; if the event occurs, we set $T_E = T_O + 1$. 

Our goal is to efficiently estimate $\mathbb{P}_{X\sim P}(T_O(X)< T_E(X))$. To this end, although we will use predictive performance as a metric to measure equivalent performance at different levels of sampling or computational budget, our empirical analysis is limited to analyzing how efficient each method is at converting next token inference to an estimate of the above probability. As a result, this generative prediction method will not be directly compared to other outcome prediction methods. See Appendix~\ref{apdnom} for collected notation.

\FloatBarrier

\section{Description of SCOPE and REACH}

\begin{figure}[htb]
    \centering
    \includegraphics[width=\linewidth]{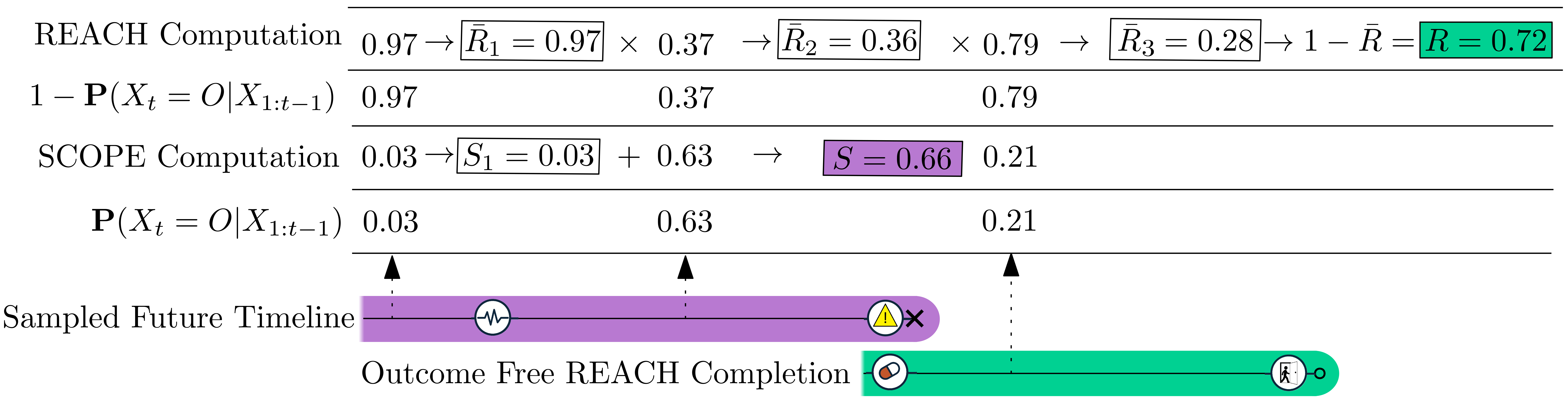}
    \caption{Step by step example computation of SCOPE and REACH. SCOPE is calculated by summing the probability that the next token is the outcome of interest at each token until the outcome of interest, marked (!), appears. Using the timeline the full timeline before this outcome of interest, an outcome free completion of the timeline is sampled. REACH is computed as 1-$\prod(1-p_i)$ where $p_i$ is the probability that the outcome of interest token would be token $i$.}
    \label{fig:s_v_r}
\end{figure}

\subsection{Sum of Conditional Outcome Probability Estimates (SCOPE)}
The Sum of Conditional Outcome Probability Estimates method, referred to as $\mathcal{S}$ in equations, begins with the same process as the Monte Carlo method: our model generates a pool of potential future timelines. However, where the above method only uses the next token distribution to sample, this method incorporates additional information from this distribution by taking a sum of the probabilities that the next token is the outcome of interest token to compute an unbiased estimator. This sum terminates at either the end of the timeline or when the outcome of interest token is selected as the next token. By saving the next token probabilities of all relevant outcomes, an often negligible memory cost, this estimator can be computed using the same pool of reusable sampled timelines as the Monte Carlo method.

Since each sub-estimator is a sum of next token probabilities, the estimator produces probabilities with much higher resolution than those of the Monte Carlo estimator. However, since the values for this estimator are a sum of raw probabilities, there are situations where the estimator can produce a value above $1$. In situations where this is more probable, the variance of this estimator can exceed the variance of the MC method. Additionally, this introduces the following double bind. If SCOPE predictions are not clipped to fall in the $[0,1]$ range, they are not valid probability estimates. However, if they are clipped to this range, SCOPE is no longer guaranteed to be an unbiased estimator. 

\subsection{Risk Estimation from Anticipated Conditional Hazards (REACH)}
Unlike the two above methods, Risk Estimation from Anticipated Conditional Hazards, which is referred to as $\mathcal{R}$ in equations, samples timelines where the outcome of interest token is \textbf{excluded} from the pool of possible next token candidates. These timelines are usually generated by setting the next token probability for the outcome to zero and re-normalizing the next token distribution, while simultaneously tracking what the probability of this outcome would have been had it not been zeroed out. After this timeline has been generated, the generated timeline is then traversed once more from the first new token. At each step, the patient has a chance of experiencing the outcome of interest with probability equal to the original next token probability of the outcome of interest. It is shown in Appendix~\ref{app:eq_prob} that this two step method results in the patient experiencing the outcome of interest with equal probability to the standard generation method. Most critically, since we know the probabilities at each step for the second part of this process, we can compute the probability of the outcome of interest given an outcome free timeline explicitly rather than through sampling, integrating out all variance created by determining whether or not the outcome of interest actually occurs. REACH can formally be considered a Rao-Blackwellization in this alternative two-step probability space (Appendix \ref{app:re_rao}). See~\figurename{}~\ref{fig:s_v_r} for a visual comparison of SCOPE and REACH.

As a result, REACH is proven to be a lower variance unbiased estimator than both of the techniques described above. The primary limitation of this method is that, unlike Monte Carlo and SCOPE, REACH requires some additional inference to be performed on a task by task basis. Since the outcome of interest is prohibited at each step in REACH timelines, starting from a pool of normally generated timelines, REACH will require all timelines that contain the outcome of interest to be re-completed starting just before the outcome of interest first appears, implying that the marginal cost of running REACH on additional outcomes is proportional to the probability of the outcome of interest. Fortunately, this implies that the outcome-marginal computational cost of REACH is lowest for rarer outcomes, where REACH offers the greatest potential reductions in variance. 

\section{Related Work}
Existing statistical methods, namely importance sampling and splitting, could also help estimate this probability \citep{kleijnen2010variance}. In essence, importance sampling techniques modify the underlying sampling distribution in order to increase the probability of sampling some outcome. Then, weights are applied when aggregating samples to ensure that the final estimator is unbiased. This technique could be applied to our problem by scaling up the probability of sampling the outcome of interest at each step and then weighting accordingly. However, since importance sampling alters the underlying probability distribution according to the outcome of interest, it would require timelines to be sampled separately for each predictive task, undermining the primary advantage of the generative prediction method. Another common method, splitting, which assigns risk scores to partially generated sequences and then re-samples these partially generated sequences for completion with higher probability of sampling high risk partial sequences, similarly requires selecting a single outcome of interest.

While potentially powerful for single outcome prediction, these approaches undermine the key appeal of generative foundation model based outcome prediction: the ability to predict an arbitrary number of outcomes at little to no marginal cost. As a result, a key consideration is to ensure that our methods remain as efficient as possible in predicting an arbitrary number of outcomes. 

Notably, SCOPE avoids these issues entirely by using the same pool of sampled future timelines to predict for an arbitrary number of outcomes. REACH, on the other hand, is able to re-use all sampled tokens that occur prior to the generation of the outcome of interest, meaning the additional token generation required is proportional to the probability of the outcome of interest. Additionally, it is shown in Appendix \ref{app:reach_wins} that REACH is in fact a Rao-Blackwellization, and thus a variance reduction, of any naive importance sampling scheme. 

Conditioning and other variance-reduction techniques have a rich history in Monte Carlo estimation. See Asmussen \& Glynn~\cite{asmussen_stochastic_2007} for an introduction. REACH can be viewed as importance sampling applied to SCOPE (see Appendix~\ref{s:var_less}) with proposal distribution $\hat p$. In the setting of Markov chains, McKeague and Wefelmeyer~\cite{McK00} introduced an estimator similar to SCOPE and showed that, for reversible Markov chains, it improves asymptotic variance. Previous research indicates that token sequences derived from EHRs tend to be highly non-Markovian, however, with long-range dependencies, interspersed repeating tokens, and increasing perplexity~\cite{Wor25}. It should also be remarked that SCOPE is neither a form of conditional Monte Carlo nor a Rao-Blackwellization, as there are cases in which it does not reduce variance (see Appendix~\ref{app:counter} for one such example). In contemporaneous work, Amar et al.~\citep{amar_integrating_2025} introduce path-computing probabilities that score the probability of a target token along a sequence. Their approach is reminiscent of REACH; however the authors do not appear to use an outcome-avoiding distribution or consider statistical properties of their estimator.

\section{Proofs}
\subsection{Derivation and Unbiasedness of \texorpdfstring{$\mathcal{S}$}{S} (SCOPE)}
Consider that $\mathbbm{1}_{\{T_O(X) < T_E(X)\}}$, the indicator random variable (r.v.) describing the event where the outcome of interest comes before the end of time timeline, is equal to 
$\textstyle\sum_{t =1}^\infty\mathbbm{1}_{\{T_O(X) = t\}}\mathbbm{1}_{\{T_E(X) > t-1\}}$, the sum of indicator random variables across each token of X describing the event where the outcome of interest token first appears at index $t$ and before the end of the timeline. The expectation of the above sum of random variables, and therefore also the probability that the outcome of interest occurs before the end of the timeline, is shown to equal the expectation of $\sum_{t=1}^{\min\{T_E(X), T_O(X)\}} \mathbb{P}(X_t =O | X_{1:t-1})$. For each each token prior to both the outcome of interest and the end of the timeline, if $X_t$ is the outcome of interest, which occurs with probability $p=\mathbb{P}(X_t =O | X_{1:t-1})$, $p$ then undercounts the real value of the indicator r.v. by $1-p$. If $X_t$ is not the outcome of interest, $p$ overcounts the correct value of the r.v. by $p$ with probability $1-p$. These over and undercounting effects therefore cancel in expectation, demonstrating that this sum provides an unbiased estimator of the probability that the outcome token will be generated before the end of the timeline. See Appendix~\ref{s:scope_derivation} for details.

\subsection{Characterization of the Variance of \texorpdfstring{$\mathcal{S}$}{S}}
By the fact that each sample is i.i.d., $\Var(\mathcal{S}) = \Var(S) /n$.
Therefore by the variance of a Monte Carlo estimator, letting $\mu = \mathbb{E}(\mathcal{S}) = \mathbb{E}(M_0)$:
{\smaller\begin{align*}
    &\Var(\mathcal{S}) - \Var(M_0) = \frac{\Var(S)- \mu(1 - \mu)}{n} = \frac{\mathbb{E}(S^2) - \mu^2- \mu + \mu^2}{n}= \frac{1}{n}\mathbb{E}\big[S(S-1)\big]\\
    &\Var(\mathcal{S}) < \Var(M_0) \iff \mathbb{E}\big[S(S-1)\big] < 0
\end{align*}}
Note that if $S < 1$ with probability $1$ then this statement is always true. However, for any timeline where there are sufficiently many tokens where the probability of the outcome of interest is sufficiently high, $S$ will indeed exceed $1$. This means that $\mathcal{S}$ is not necessarily a lower variance estimator than $M_0$ and should therefore be checked as an alternative for $M_0$ rather than a universal replacement. Additionally, since $S$, as an unbiased estimator for $\mu$, is more likely to be large for more probable events, $\mathcal{S}$ is often a higher variance estimator than $M_0$ in contexts where the outcome of interest is more likely. This fact is reflected in our synthetic and empirical results. However, as is shown in Appendix \ref{app:counter}, $\Var(\mathcal{S}) > \Var(M_0)$ can hold for arbitrarily low mean. Fortunately, the additional computational cost required to compute both $M_0$ and $\mathcal{S}$ for a given outcome is limited to saving and summing the probabilities, which is negligible when compared to the computational cost of inference.

\subsection{Derivation and Unbiasedness of \texorpdfstring{$\mathcal{R}$}{R} (REACH)}
Moving to a study of the second new estimator, $\mathcal{R}$, let the event $A$ be the subset of all timelines that, at some point before the end of the timeline, contain some outcome of interest token $O$. From this, we can define a Monte Carlo estimator for $P(A)$ by sampling $m$ timelines and using the fraction of samples that include the outcome of interest token as an unbiased estimate of $P(A)$.

Define an outcome-avoiding probability space $\Omega_{\not O}$ as follows. Timelines are generated using the same model and time limit as in the previous probability space; however, in this probability space, the probability of the next token being the outcome of interest token is set to zero and the other probabilities are re-normalized. Additionally, for each new token in our potential future timeline, we perform a Bernoulli trial with probability of success equal to the probability that the outcome of interest token would have been generated at this token under the original paradigm. 

Let $B$ be the subset of the probability space where at least one of the Bernoulli trials along the timeline is successful. We show $P(A) = P(B)$ in Appendix \ref{app:eq_prob} by constructing the natural bijection between the complements $A^\complement$ and $B^\complement$ where each timeline in $A^\complement$, which must by definition be outcome-free, is mapped to the element in $B^\complement$ with an identical timeline and a series of Bernoulli trials where all are failures. This bijection preserves probabilities so that $P(A) = P(B)$.

By the tower rule, $P(B) = \mathbb{E}(\mathbbm{1}_B) = \mathbb{E}_{\hat X_{1:n}}(\mathbb{E}(\mathbbm{1}_B | \hat X_{1:n}))$, so $\mathbb{E}(\mathbbm{1}_B | \hat X_{1:n})$ is an unbiased estimator for $P(A)$ for any sampled timeline $\hat X_{1:n}$. Since $\mathbbm{1}_B | \hat X_{1:n}$ is simply the event that at least one Bernoulli trial succeeds, it is clear that
{\smaller\[
    \mathbb{E}(\mathbbm{1}_B | \hat X_{1:n}) = 1 - \textstyle\prod_{t = 1}^{T_E}(1 -  P(X_t = O |\hat X_{1:t-1}))
\]}

which reveals that our estimator $\mathcal{R}$ is therefore an unbiased estimator for the probability that a sampled timeline will contain the outcome of interest token. 

\subsection{Characterization of the Variance of \texorpdfstring{$\mathcal{R}$}{R}}
Letting $m_0$ be the samples that determine $M_0$ and $R$ be the samples that determine $\mathcal{R}$, by the law of total variance:
{\smaller
\begin{align*}
    \Var(M_0) &=\frac{1}{n}\Var(m_0) = \frac{1}{n}\Var(\mathbbm{1}_A)= \frac{1}{n}\Var(\mathbbm{1}_B) = \frac{1}{n}\big({\Var(\color{reach}\mathbb{E}(\mathbbm{1}_B | \hat X_{1:n})}) + \mathbb{E}(\Var(\mathbbm{1}_B | \hat X_{1:n}))\big)\\
    &= {\frac{1}{n}\big(\Var(\color{reach} R}) + \mathbb{E}(\Var(\mathbbm{1}_B | \hat X_{1:n}))\big)  
    = \Var(\mathcal{R}) + \mathbb{E}(\Var(\mathbbm{1}_B | \hat  X_{1:n}))/n
\end{align*}}
so that:
\begin{equation}
    \Var(M_0) - \Var(\mathcal{R}) = \mathbb{E}(\Var(\mathbbm{1}_B | \hat X_{1:n}))/n.
    \label{eq:spont}
\end{equation}
This also implies that $\Var(\mathcal{R}) \leq \Var(M_0)$.  Intuitively, the above means that $\mathcal{R}$ presents a larger improvement to $M_0$ in scenarios in which an outcome-free timeline has a higher degree of uncertainty regarding the outcome of interest, with the difference becoming larger the closer $p$ is to $0.5$.  

We call this statistic, $\mathbb{E}(\Var(\mathbbm{1}_B | \hat X_{1:n})),$ ``\textbf{spontaneity}.'' In Appendix \ref{app:spont_prev}, it is shown that the efficiency gains presented by SCOPE and REACH can be predicted by a linear model using only prevalence and spontaneity as inputs with an $R^2$ of over 0.93. The relationship between spontaneity and estimator variance is further explored in synthetic Markov chain experiments described in Appendices \ref{app:mc_methods} and\ref{app:mc_synthetic}.

In addition to being a lower variance estimator than the Monte Carlo method, $\mathcal{R}$ is also a lower variance estimator than $\mathcal{S}$ for all models and stopping conditions. This fact is proven in detail in Appendix \ref{s:var_less}. 

\subsection{Empirical Study Methods}
In addition to theoretical guarantees, it is important to study how these methods perform on real world patient data. We generated risk predictions for 11 clinically important outcomes across MIMIC-IV~\citep{johnson_mimic-iv_2023} and an EHR data source from the University of Chicago Medical Center (UCMC), of which only 10 of outcomes were available in the UCMC data. Both datasets are converted to the Common Longitudinal ICU Format \citep[CLIF:][]{Roj25} version 2.1 standard and tokenized (Appendix \ref{app:tokenize}) using our 1385-token vocabulary. Timelines are split temporally, $70\%$ for training, $10\%$ for tuning, and $20\%$ held out for evaluation. The training split is then used to train a modified version of the Llama-3.2-1B model on the self-supervised task of next token prediction \citep{dubey2024llama3herdmodels}. 

From the held out timelines, we remove stays in which the patient is discharged within the first 24 hours. Remaining timelines are truncated to tokens occurring in the first 24 hours of the stay and passed to the trained model for inference. We generate 100 potential future timelines per patient using SGLang's inference engine \citep{sglang2024arxiv}, with the model auto-regressively generating tokens until the patient is discharged or the maximum timeline length of 10,000 tokens is reached. SCOPE and REACH are computed iteratively from the next token distribution during inference. When a timeline generates an outcome of interest, the corresponding outcome-free completion required by REACH is generated in the same batch to avoid KV-cache misses. Confidence intervals are obtained by bootstrap resampling 50\% of the eligible validation set 100 times.

To compute a concrete inference cost reduction factor, we proceed as follows. For each bootstrap and outcome, we take the $100$-timeline Monte Carlo estimator as a baseline. Given a tolerance $\epsilon$ (set to $0.01$ here), we find for each estimator the smallest sample size $n$ such that AUROC remains within $\epsilon$ of the baseline for all sample sizes greater than $n$. To ensure fair comparison, this statistic is also computed for Monte Carlo itself, and savings are reported as $n_{MC}/n_{\text{new}}$. For example, if Monte Carlo remains within $\epsilon$ of $100$-sample accuracy down to $60$ samples while SCOPE does so down to $15$, this is reported as a $4\times$ reduction. The same definition extends to token efficiency by letting $n$ count generated tokens rather than sampled timelines.

This framework does not capture generative efficiency when multiple outcomes are estimated simultaneously. We therefore extend the per-task $\epsilon$ token cost to a cross-outcome total $\epsilon$ cost: the inference required to stay within $\epsilon$ of $100$-sample Monte Carlo on all tasks. For Monte Carlo and SCOPE, cost is shared across tasks and equals the maximum individual task cost. REACH must still generate at least as many standard timelines as the most demanding outcome, but for less demanding outcomes it can subsample these timelines and perform completions only on the subsampled timelines containing the outcome of interest, slowing the rate at which its cost grows with the number of outcomes. To examine the total cost across all three methods, for each $k\in[11]$, we report the mean cost across all possible sets of $k$ outcomes.

The necessary code to run these experiments, as well as the config files required to use the \href{https://github.com/bbj-lab/cocoa}{cocoa} and \href{https://github.com/bbj-lab/cotorra}{cotorra} libraries to tokenize the dataset and train the initial model, are made available \href{https://github.com/lukesolo-ml/SCOPE-REACH-Codebase-Companion}{here}.
\section{Results}
\subsection{Experimental Results}
\begin{figure}[th!]
\centering
\includegraphics[width=\textwidth]{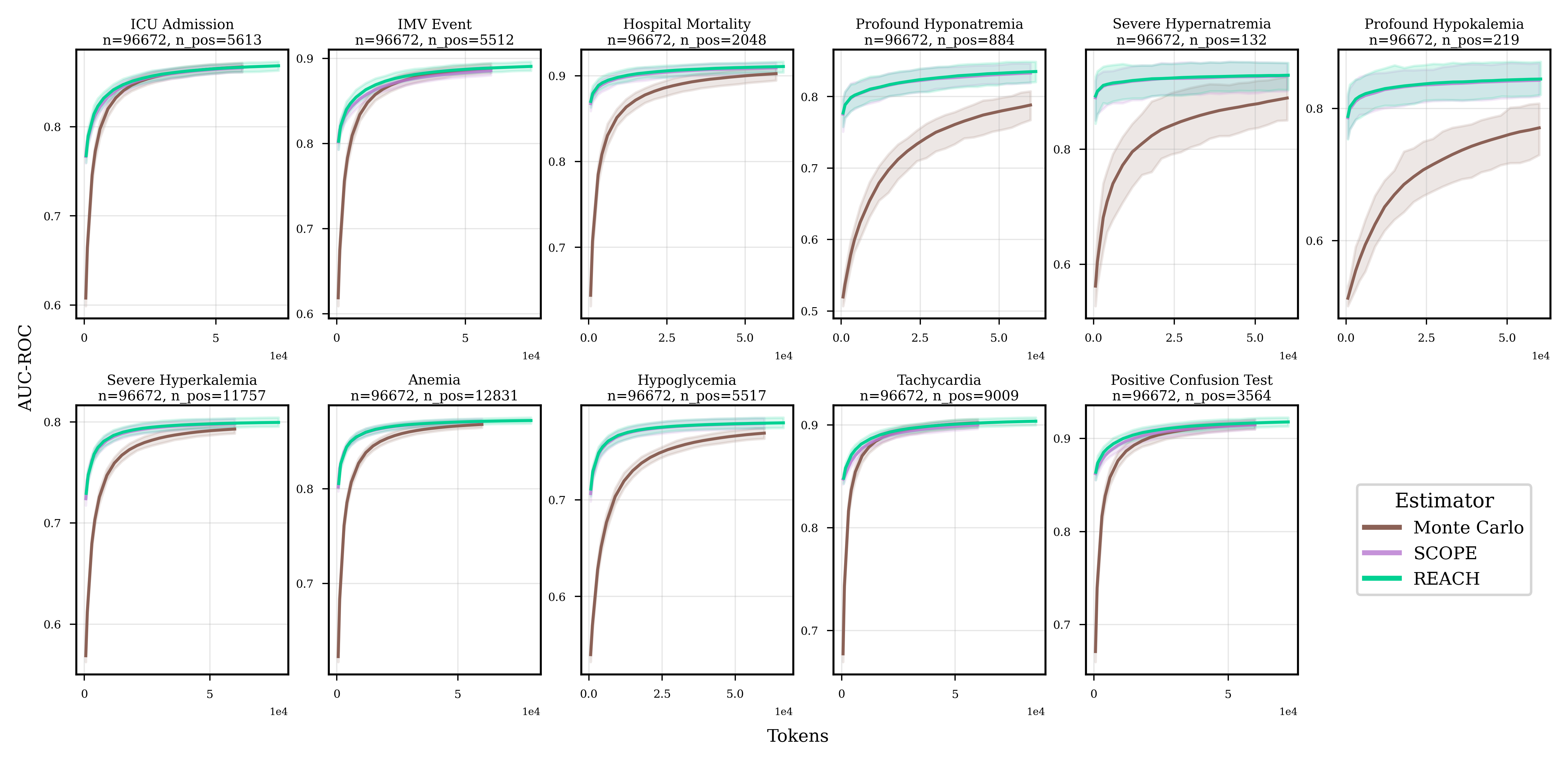}
\caption{Prediction accuracy comparison across MIMIC-IV patients on all $11$ outcomes. The x-axis tracks the number of tokens, in increments of 10,000 generated per patient to achieve the predictive accuracy reflected on the y-axis. UCMC data curves can be found in Appendix \ref{app:ucmc_auc}.}
\label{fig:convergence}
\end{figure}
\figurename{}~\ref{fig:convergence} plots AUROC against tokens generated per patient for all 11 MIMIC-IV outcomes. SCOPE and REACH match 100-sample Monte Carlo performance with substantially fewer tokens on every outcome except ICU Admission, and the gap is largest for rare lab-defined outcomes (severe hypernatremia, profound hypokalemia, profound hyponatremia, severe hyperkalemia, hypoglycemia), where the Monte Carlo curve remains below the asymptote even at the full 100 timelines per patient budget. Calibration is preserved across estimators (per-outcome curves and Brier scores in Appendices \ref{app:curves_and_cal_1} and \ref{app:curves_and_cal_2}).

\tablename{}~\ref{tab:token_efficiency_combined} quantifies these savings at $\epsilon = 0.01$. The most extreme reductions occur for severe hypernatremia and profound hypokalemia (85.00× for SCOPE and 83× for REACH on MIMIC-IV); profound hypokalemia on UCMC shows comparable 75.00× and 72.89× reductions. Aggregate means are 20.97× (SCOPE) and 20.71× (REACH) on MIMIC, and 11.10× and 11.00× on UCMC, while the medians of 2.50× to 3.41× better reflect typical outcomes, since the means are pulled upward by the rare-event tail. The rank ordering of outcomes by efficiency gain replicates across institutions, supporting external validity.
 
\begin{table}[tb]
\centering
\caption{Single task token efficiency of SCOPE and REACH vs.\ Monte Carlo baseline at $\varepsilon=0.01$, on MIMIC-IV and UCMC. Values are mean token cost per patient. REACH cost for each outcome reflects only the cost of generating REACH completions for said outcome. \textbf{Bold} indicates lowest token cost. Ratios $>1$ indicate fewer tokens needed than MC. (Full table with confidence intervals is available in Appendix \ref{app:full_t_1})}
\label{tab:token_efficiency_combined}
\small
\setlength{\tabcolsep}{3pt}
\begin{tabular}{@{}lrrr cc rrr cc@{}}
\toprule
& \multicolumn{5}{c}{\textbf{MIMIC-IV}} & \multicolumn{5}{c}{\textbf{UCMC}} \\
\cmidrule(lr){2-6} \cmidrule(lr){7-11}
\textbf{Outcome}
  & \boldmath$T_{\text{MC}}$ & \boldmath$T_{\text{SC}}$ & \boldmath$T_{\text{RE}}$
  & \textbf{MC/SC} & \textbf{MC/RE}
  & \boldmath$T_{\text{MC}}$ & \boldmath$T_{\text{SC}}$ & \boldmath$T_{\text{RE}}$
  & \textbf{MC/SC} & \textbf{MC/RE} \\
\midrule
ICU Admission
  & \textbf{30k} & \textbf{30k} & 31k & 1.00 & 0.95
  & 73k & 55k & \textbf{53k} & 1.33 & 1.39 \\
IMV Event
  & 33k & 36k & \textbf{26k} & 0.92 & 1.25
  & 73k & 92k & \textbf{66k} & 0.80 & 1.11 \\
Hospital Mortality
  & 33k & \textbf{5993} & 6413 & 5.50 & 5.14
  & 101k & \textbf{28k} & 29k & 3.67 & 3.54 \\
Profound Hyponatremia
  & 51k & \textbf{1199} & 1240 & 42.50 & 41.09
  & 119k & \textbf{28k} & 29k & 4.33 & 4.16 \\
Severe Hypernatremia
  & 51k & \textbf{599} & 611 & 85.00 & 83.31
  & 92k & \textbf{9181} & 9670 & 10.00 & 9.49 \\
Profound Hypokalemia
  & 51k & \textbf{599} & 613 & 85.00 & 83.16
  & 138k & \textbf{1836} & 1889 & 75.00 & 72.89 \\
Severe Hyperkalemia
  & 30k & 12k & \textbf{11k} & 2.50 & 2.68
  & 73k & 28k & \textbf{22k} & 2.67 & 3.27 \\
Anemia
  & 27k & 12k & \textbf{11k} & 2.25 & 2.42
  & 55k & 28k & \textbf{24k} & 2.00 & 2.30 \\
Hypoglycemia
  & 36k & 8991 & \textbf{6788} & 4.00 & 5.30
  & 92k & \textbf{9181} & 10k & 10.00 & 8.91 \\
Tachycardia
  & 24k & 27k & \textbf{21k} & 0.89 & 1.12
  & 55k & 46k & \textbf{19k} & 1.20 & 2.92 \\
Positive Confusion
  & 27k & 24k & \textbf{20k} & 1.12 & 1.36
  & --- & --- & --- & --- & --- \\
\midrule
\multicolumn{1}{@{}l}{\textit{Aggregate:}}
  & & & & \textbf{SC} & \textbf{RE}
  & & & & \textbf{SC} & \textbf{RE} \\
\quad Mean
  & & & & 20.97 & 20.71
  & & & & 11.10 & 11.00 \\
\quad Median
  & & & & 2.50 & 2.68
  & & & & 3.17 & 3.41 \\
\bottomrule
\end{tabular}
\end{table}

\begin{figure}[b]
  \centering
  \begin{subfigure}[t]{0.49\linewidth}
    \centering
    \includegraphics[width=\linewidth]{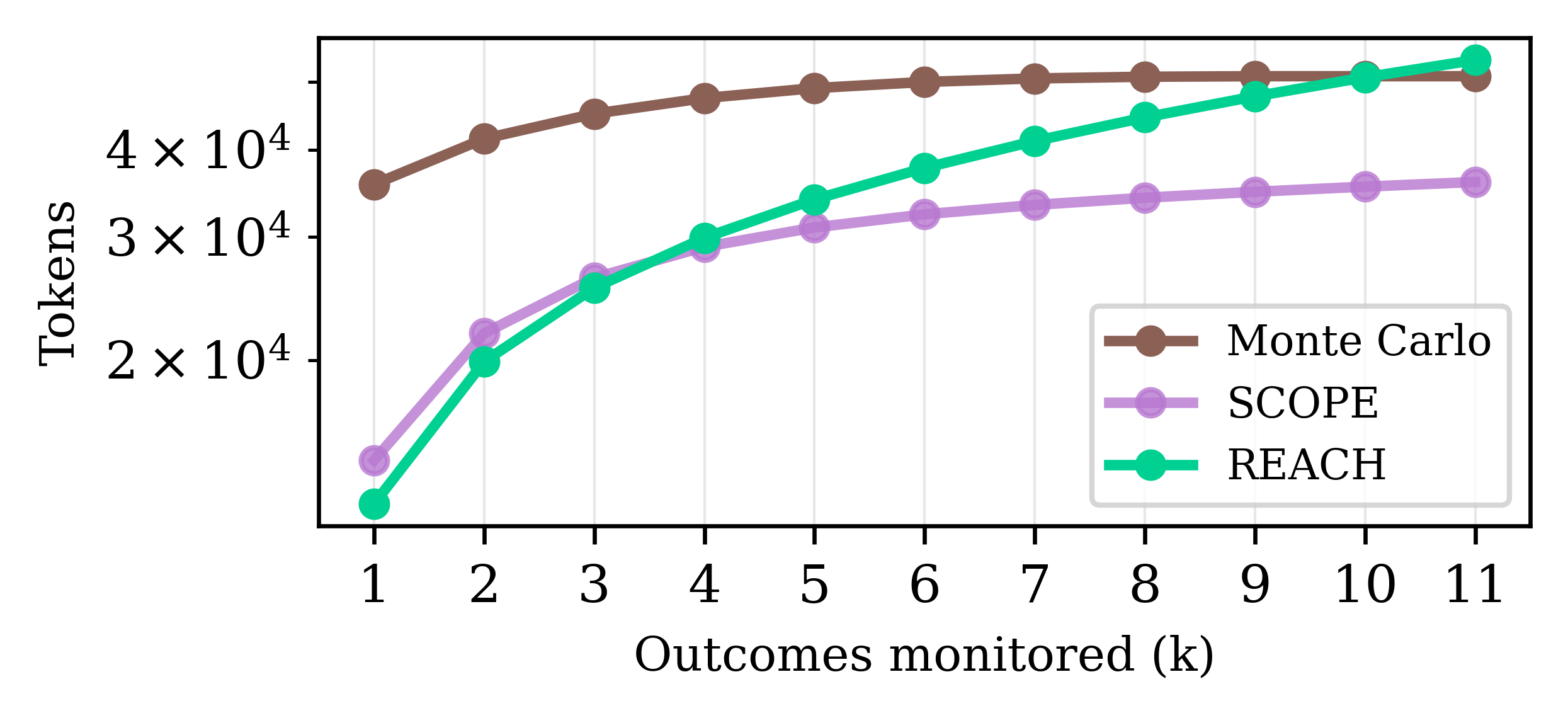}
    \caption{MIMIC}
    \label{fig:cost:a}
  \end{subfigure}
  \hfill
  \begin{subfigure}[t]{0.49\linewidth}
    \centering
    \includegraphics[width=\linewidth]{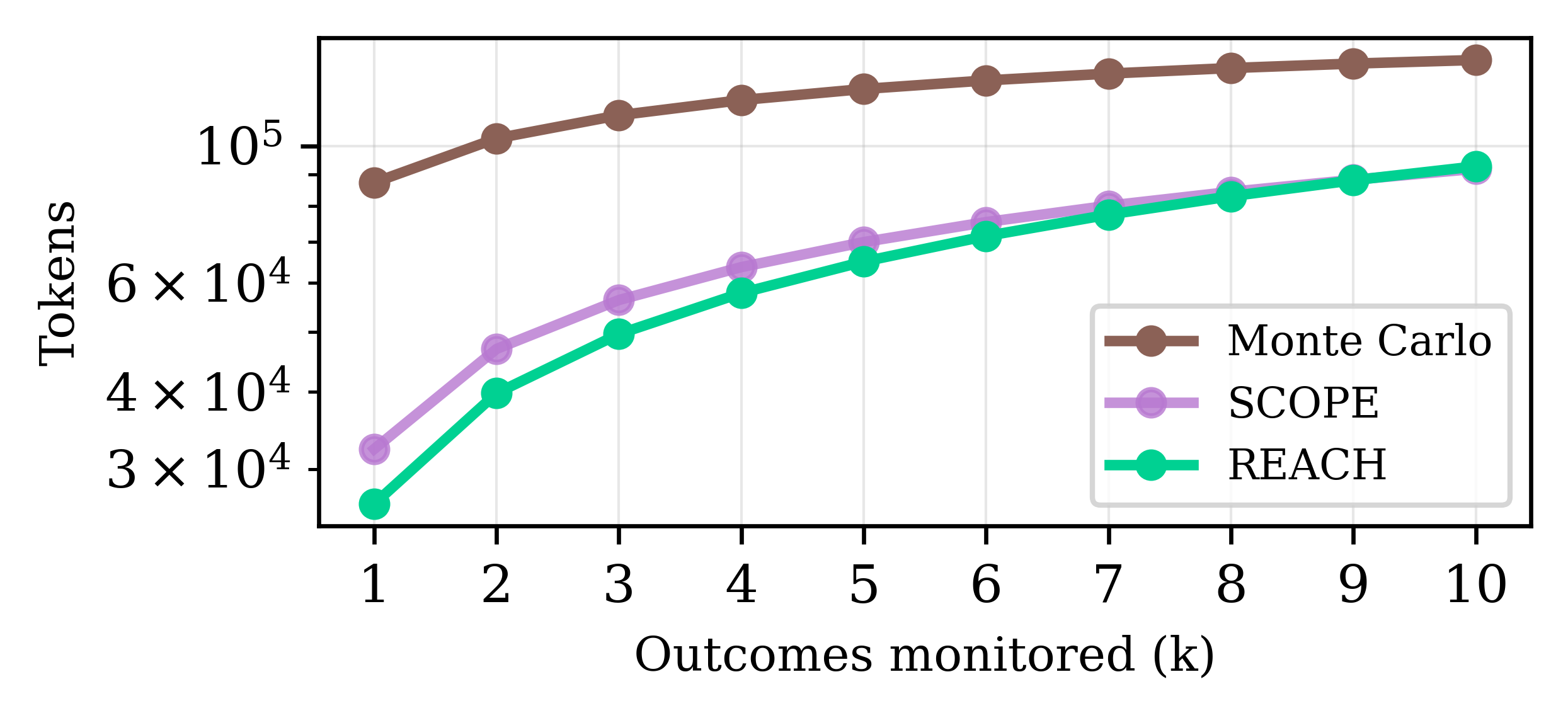}
    \caption{UCMC}
    \label{fig:cost:b}
  \end{subfigure}
  \caption{Mean cost per patient across all size $k$ outcome subsets in tokens for predicting outcomes within tolerance $\epsilon=0.01$ of 100-sample accuracy for both (a) MIMIC and (b) UCMC. The $x$-axis gives
           the number of outcomes and the $y$-axis gives mean per-patient cost token
           cost.
           }
  \label{fig:token-cost}
\end{figure}

\figurename{}~\ref{fig:token-cost} shows cumulative token cost as the number of jointly monitored outcomes grows from 1 to 11 (MIMIC) or 10 (UCMC). Because SCOPE reuses a single sampled pool across all outcomes at no marginal generation cost, its cumulative cost flattens almost immediately and remains well below Monte Carlo for every k. REACH grows more steeply, since each outcome requires its own bank of outcome-free completions, but remains at or below Monte Carlo throughout, only slightly exceeding MC for the MIMIC-IV dataset with all $11$ outcomes. This pattern supports a hybrid deployment strategy: SCOPE as the default for arbitrary multi-outcome panels, with REACH reserved for the highest-priority outcomes where its variance guarantee is most valuable.
           
These results demonstrate clearly that SCOPE and REACH are capable of delivering consistent reductions in the amount of required inference for generative outcome prediction, especially for rare outcomes. Of our $11$ outcomes, only on the the task of ICU Admission do both SCOPE and REACH fail to deliver reductions in required inference. This failure mode is consistent with our spontaneity-based account: ICU admission is a clinician-initiated action that is typically signaled in advance by tokens already present in the timeline, leaving little residual outcome uncertainty for REACH to integrate out and giving SCOPE little dispersion to resolve. We recommend empirical screening with the spontaneity statistic before assuming gains on a new outcome.

\section{Conclusion}
SCOPE and REACH address the three limitations of Monte Carlo generative outcome prediction identified in the introduction: estimator dispersion, computational cost, and sampling variance. Both estimators are unbiased, REACH is provably lower variance than either Monte Carlo or SCOPE for any model and outcome, and across $11$ clinically important outcomes in two health systems they match $100$-sample Monte Carlo accuracy with median reductions of $2.5\times$ to $3.4\times$ in tokens generated and reductions exceeding $80\times$ for the rarest outcomes. The estimators are complementary in deployment: SCOPE is the natural default for multi-outcome panels, since a single sampled pool yields estimates for an arbitrary number of outcomes at no marginal generation cost, while REACH is preferred when a small number of high-priority outcomes justify task-specific completions in exchange for its variance guarantee.

The magnitude of the improvement is governed by two properties of the prediction task: prevalence and spontaneity. A linear model in $\log$-prevalence and $\log$-spontaneity explains over $93\%$ of the variance in efficiency gains across outcomes (Appendix \ref{app:spont_prev}), and the rarest, most spontaneous outcomes, those for which the patient's history describes a risk rather than a guarantee, benefit most. This is precisely the regime in which Monte Carlo's arithmetic resolution fails and in which clinical decision-making is most sensitive to small absolute differences in risk.

These properties translate directly to deployment constraints. Lower inference cost makes more frequent, lower-latency risk updates feasible in acute settings such as the ICU. Additionally, lower compute requirements for inference reduce energy usage for this existing class of models. Continuous risk estimates replace stepwise Monte Carlo proportions with smoother trajectories, separating physiologic change from sampling noise and supporting the construction of foundation-model-derived survival curves (Section \ref{ss:limitations}). Together, these contributions move generative EHR foundation models meaningfully closer to real-time clinical decision support.

\subsection{Limitations}
\label{ss:limitations}
Three limitations bound these results. First, despite consistently strong performance and reliable reductions in required inference across most tasks, SCOPE lacks rigorous guarantees of variance reduction and SCOPE estimators can exceed $1.0$. Second, as illustrated in \figurename{}~\ref{fig:token-cost}, even with rare outcomes, the task-marginal cost of REACH can add up. In applied settings, it may be prudent to only compute REACH completions for the least prevalent outcomes. Finally, both estimators require the outcome to be captured by the appearance of a single token requiring pre-processing to inject computed label tokens where multi-token logic is required.

\subsection{Future Work}
Since these estimators provide continuous risk scores for patient timelines, rather than binary labels of whether or not the outcome of interest token was observed, these methods could be used to improve clinician understanding of generative event models by identifying common events or features of high risk timelines. Furthermore, not only are these risk scores continuous at the whole timeline level, but for REACH in particular, the risk scores evolve over time, which could provide clinicians with clearer intuition regarding the trajectory of a patient with respect to specific outcomes. This potential for the construction of foundation model derived survival curves has a large number of potentially valuable applications. 

Additionally, while these methods improve the computational cost of generative risk prediction by resolving the dispersion of estimates and reducing variance of estimators, future work could further address this problem by varying the number of sampled timelines per patient. This would allow this technique to expend fewer computational resources computing the risk in situations where the outcome of interest is either impossible or guaranteed, allowing for compute time to be focused on risk stratification in more ambiguous cases. 

\subsection*{Acknowledgments}
This work was supported in part by the National Institutes of Health, specifically the National Institute of Neurological Disorders and Stroke grant R00NS114850 to BKB. This project would not have been possible without the support of the Center for Research Informatics at the University of Chicago and particularly the High-Performance Computing team led by Mike Jarsulic. The authors are grateful for the resources and support this team provided throughout the duration of the project. The Center for Research Informatics is funded by the Biological Sciences Division at the University of Chicago with additional funding provided by the Institute for Translational Medicine/CTSA (NIH UL1TR002389).

\bibliographystyle{unsrt}
\bibliography{neurips_2026}


\appendix
\clearpage
\section{Nomenclature}
\label{apdnom}
\begin{description}
\item[$M_0$] Equation shorthand for the Monte Carlo estimator.
\item[$\mathcal{S}$] Equation shorthand for the SCOPE estimator.
\item[$\mathcal{R}$] Equation shorthand for the REACH estimator.
\item[$T_O$] The random variable equal to the index of the first outcome of interest token.
\item[$T_E$] The random variable equal to the index of the terminating token of a sequence or the token coming after the first outcome of interest token.
\item[$A$] The event describing the subset of timelines in the first probability space that contain the outcome of interest token.
\item[$B$] The event describing the subset of timelines that have a successful Bernoulli trial under the second probability space. 
\item[$V$] The vocabulary of tokens.
\item[$O$] The outcome of interest token, an element of $V$.
\item[$X_{1:n}$] A tokenized timeline of length $n$.
\item[$\hat X_{1:n}$] A tokenized timeline of length $n$ that excludes $O$ from the set of possible next tokens. 
\item[$\Omega_{M_\tau}$] The probability space of timelines sampled under some model $M$ and time limit $\tau$.
\item[$\Omega_{\not\mathcal{O}}$] The probability space of outcome-free timelines and Bernoulli trials at each token.
\item[$\textbf{c}_{1:n}$] The result of $n$ Bernoulli trials for some outcome-free tokenized timeline of length $n$.
\end{description}

\newpage
\section{Derivation of SCOPE}
\label{s:scope_derivation}
\begin{align*}
    \mathbb{P}(T_O < T_E) = \mathbb{E}_{X\sim P}[\mathbbm{1}_{T_O(X) < T_E(X)}]= \mathbb{E}_{X\sim P}\big[\textstyle\sum_{t =1}^\infty\mathbbm{1}_{\{T_O(X) = t\}}\mathbbm{1}_{\{T_E(X) > t-1\}} \big]
\end{align*}
Equivalent to the product of the indicator r.v.s in the interior of the term above, to simplify the notation, define a new function $f$ over partial sequences: 
{\smaller\[
g(x_{1:t-1}) =  \mathbbm{1}_{\{O \notin x_{1:t-1}\}} \mathbbm{1}_{\{T_E( x_{1:t-1}) > t\}}, \quad f(x_{1:t}) =  \mathbbm{1}_{\{x_t = O\}}g(x_{1:t-1})
\]}
 
Note that for any sequences exceeding the time horizon or containing $O$, the above function will map to zero. 
So that, by the linearity of expectation and the law of total expectation:
{\smaller \begin{align*}
    \mathbb{P}( T_O < T_E )
    & = \textstyle \sum_{t=1}^{\infty} \EE_{X\sim P} \big[ f(X_{1:t}) \big]=  \textstyle \sum_{t=1}^{\infty} \EE_{X_{1:t-1}\sim P} \big[ \EE[f(X_{1:t}) | X_{1:t-1} ]\big]\\
    &=  \textstyle \sum_{t=1}^{\infty} \EE_{X_{1:t-1}\sim P} \big[g(X_{1:t-1}) \EE[\mathbbm{1}_{\{X_t = O\}} | X_{1:t-1} ]\big]\\
    &=  \textstyle \sum_{t=1}^{\infty} \EE_{X_{1:t-1}\sim P} \big[g(X_{1:t-1}) \mathbb{P}(X_t = O | X_{1:t-1} )\big]
\end{align*}
\normalsize Next by the law of total expectation }
{\smaller\begin{align*}
\mathbb{E}_X(g(X_{1:t-1})\mathbb{P}(X_t =O | X_{1:t-1})| X_{1:t-1})) &= \mathbb{E}_{X_{1:t-1}}[\mathbb{E}[g(X_{1:t-1})\mathbb{P}(X_t =O | X_{1:t-1})|X_{1:t-1}]]\\
&=\mathbb{E}_{X_{1:t-1}}[g(X_{1:t-1})\mathbb{P}(X_t =O | X_{1:t-1})]
\end{align*}}
Substituting this into the above
{\smaller\begin{align*}
P(T_O < T_E) & =  \textstyle \sum_{t=1}^{\infty} \EE_{X\sim P} \big[ g(X_{1:t-1})\mathbb{P}(X_t =O | X_{1:t-1})\big] =  \EE_{X\sim P} \big[ \textstyle \sum_{t=1}^{\infty} g(X_{1:t-1})\mathbb{P}(X_t =O | X_{1:t-1})\big]\\
    & = \EE_{X\sim P} \big[ \sum_{t=1}^{\min\{T_E(X), T_O(X)\}} \mathbb{P}(X_t =O | X_{1:t-1})\big]
\end{align*}}

The inside of this expression, which we will denote $S_i$, can then be sampled to generate an unbiased estimator, $\mathcal{S} = \frac{1}{n}\sum_{i=1}^nS_i$.
\newpage

\section{Proof of Equal Probabilities}
\label{app:eq_prob}
Consider a sequence generation model $M$ that generates a probability distribution for the next token from some vocabulary $V$, which contains our outcome of interest token $O$. Let $Y_{1:n}$ denote the sequence of tokens that the model has processed before generating any new tokens. Note that each token that is associated with an increasing time stamp $\mathcal{T}_i$. Let $P_M(Y_0=t_j)$ denote the probability that the next token generated by the model is $t_j$. Likewise, if the model has already generated $Y_{1:i}$, $P_M(Y_{i+1} = t_j | Y_{1:i})$ denotes the probability that the next token generated will be $t_j$.

Before introducing our improved sampling method, we will introduce the standard Monte Carlo method. Define probability space $\Omega_{M_\tau}$:
\begin{equation}
    \Omega_{M_\tau} := \big\{X_{1:n} | \ \forall i \in [n], \ X_i \in V,\ \ \mathcal{T}_i <\tau \big\}
\end{equation}
\begin{equation}
    P_{\Omega_{M_\tau}}(X_{1:n}) = P_M(Y_0 = X_0)\prod_{i = 1}^{n}P_M(Y_i = X_i | Y_{1:i-1} = X_{1:i-1}) \label{eq:p_space1}
\end{equation}
Intuitively, the samples in this probability space correspond to the sequences of tokens generated by by sampling the tokens directly at the rate dictated by the model.
Within this distribution, define the event $A$:
\begin{equation}
    A := \{X_{1:n}\in \Omega_{M_\tau}| \ \exists i \in [n] \text{ s.t. } X_i = O \}
\end{equation}
Next, consider a second probability space $\Omega_{\not\mathcal{O}}$:

\begin{equation}
    \Omega_{\not\mathcal{O}} := \big\{(X_{1:n}, \textbf{c}_{1:n}) | \ \forall i \in [n], \ X_i \in V\setminus O,\ \ \mathcal{T}_i <\tau,\ \textbf{c}_i \in \{0,1\}] \big\}
\end{equation}
\begin{equation}
    P_{\Omega_{\not\mathcal{O}}}(X_{1:n}) =\Bigg( \dfrac{P_M(Y_0 =X_0)}{P_M(Y_0 \not= O)}\prod_{i = 1}^{n}\dfrac{P_M(Y_i = X_i | Y_{1:i-1} = X_{1:i-1})}{P_M(Y_i \not = O|Y_{1:i-1} = X_{1:i-1}) }\Bigg) \label{eq:p_space2_1}
\end{equation}
Intuitively, this can be understood as the probability of generating a given sequence $X_{1:n}$ by sampling next tokens at the rates suggested by the model but this time \textbf{excluding} the outcome token.

Define $h_i(X_{1:n}) = P_M(Y_i = O | X_{1:i-1})$

(Note that in order to keep notation simple $X_{1:n}$ may be omitted)
{\small
\begin{equation}
    P_{\Omega_{\not\mathcal{O}}}(X_{1:n}, \textbf{c}_{1:n}) = P_{\Omega_{\not\mathcal{O}}}(X_{1:n})\prod_{i = 1}^n\bigg((1-\textbf{c}_i)(1-h_i) + \textbf{c}_ih_i\bigg)  \label{eq:p_space2_2}
\end{equation}
}
Each $\textbf{c}_i$ can be understood as an independent Bernoulli trial with probability $h_i$. The purpose of these variables is to simulate the tokens in the timeline being randomly ``flipped'' to the outcome token. 

Now define the event corresponding to all outcomes where at least one token in the timeline is randomly flipped to the outcome of interest as $B$.

Consider the complements of our two events described above: $A^\complement$, which corresponds to all sampled timelines that \textbf{do not} contain the outcome of interest token, and $B^\complement$, which corresponds to all sampled timelines where \textbf{none} of the Bernoulli trials succeed and flip a token to the outcome of interest. 

Since the elements of both probability spaces are determined by the same next token distribution model, note that there exists a natural bijection $\phi:A^\complement\rightarrow B^\complement$:
\begin{equation}
    \phi(X_{1:n}) := (X_{1:n}, (0,...,0))
\end{equation}

Furthermore, from our definition of our probability spaces:
\begin{equation}
    P_{\Omega_{\not\mathcal{O}}}(\phi(X_{1:n})) = P_{\Omega_{\not\mathcal{O}}}((X_{1:n}, (0,...,))
\end{equation}

Looking at the definitions once more, it is clear that the Bernoulli trial product terms from \eqref{eq:p_space2_2} cancel with the denominator of \eqref{eq:p_space2_1} to yield exactly \eqref{eq:p_space1}. Therefore:

\begin{equation}
    P_{\Omega_{\not\mathcal{O}}}(\phi(X_{1:n})) = P_{\Omega_{M_{\tau}}}(X_{1:n})
\end{equation}

By the definition of the probability of an event, this allows for the substitution:

\begin{equation}
    P_{\Omega_{\not\mathcal{O}}}(B^C) = \sum_{(X_{1:n}, (0,...,0)) \in B^C}P_{\Omega_{\not\mathcal{O}}}((X_{1:n}, (0,...,0)))
\end{equation}
\begin{equation}
    = \sum_{X_{1:n} \in A^C}P_{\Omega_{\not\mathcal{O}}}(\phi(X_{1:n})) = \sum_{X_{1:n} \in A^C}P_{\Omega_{M_\tau}}(X_{1:n}) = P_{\Omega_{M_\tau}}(A^C) 
\end{equation}

So, by the law of total probability, we have shown that $P(A) = P(B)$.
\newpage
\section{UCMC AUC Curves}
\label{app:ucmc_auc}
\begin{figure}[h!]
    \centering
    \includegraphics[width=\linewidth]{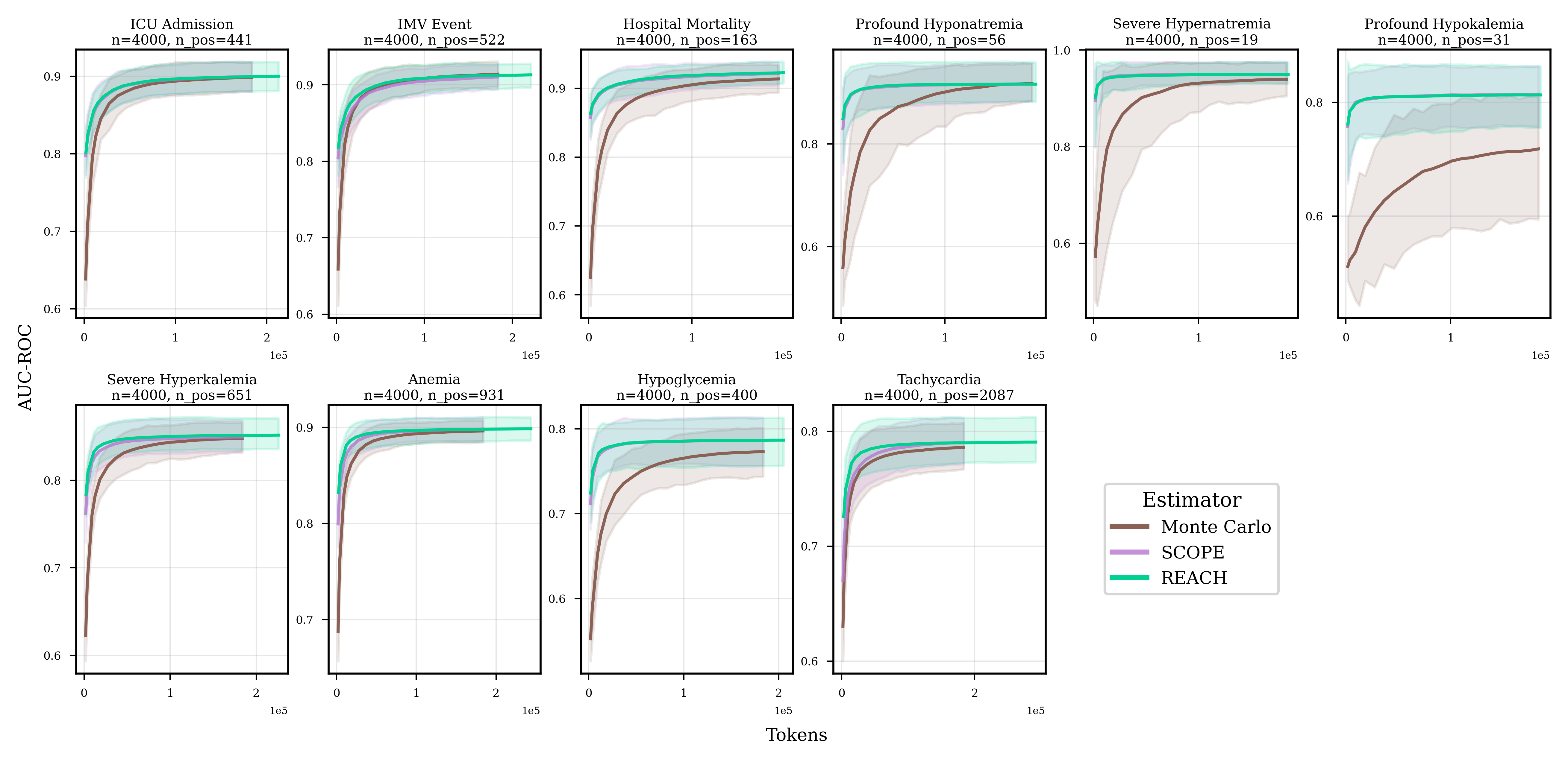}
    \caption{AUC vs Tokens Generated for UCMC}
    \label{fig:placeholder}
\end{figure}
\newpage
\section{Proof of \texorpdfstring{$\Var(\mathcal{R}) \leq \Var(\mathcal{S})$}{Var(R) <= Var(S)}}
\label{s:var_less}
From the body of the paper, recall that we can define our SCOPE estimator as 

\begin{equation}
    \mathcal{S} = \frac{1}{n}\sum_{i=1}^n\Big[\sum_{t = 1}^{\min\{T_E(X^{(i)}), T_O(X^{(i)})\}}P(X_t = O | X^{(i)}_{1:t-1})\Big]\label{eq:1}
\end{equation}

If we then define
\begin{equation}
    f_t(x) = P(X_T = O | X_{1:t-1} = x_{1:t-1})\mathbbm{1}_{\{T_O(x_{1:t-1}), T_E(x) > t-1\}}
\end{equation}
We can instead express $\mathcal{S}$ as:
\begin{equation}
    \mathcal{S} = \frac{1}{n}\sum_{i=1}^n\sum_{t\geq1}f_t(X^{(i)})\label{eq:m1}
\end{equation}

As an alternative to our previous derivation of the REACH estimator, let $\hat P$ denote the restriction of $P$ to $V\setminus \{O\}$ and apply importance sampling to \eqref{eq:1} to obtain:
\begin{equation}
    \mathcal{R} = \frac{1}{n}\sum_{i=1}^n\Big[\sum_{t = 1}^{T_E(\hat X^{(i)})}P(X_t = O | \hat X^{(i)}_{1:t-1})\frac{P(\hat X^{(i)_{1:t-1}})}{\hat P(\hat X^{(i)_{1:t-1}})}\Big]\label{eq:2}
\end{equation}
where $\forall i \in [n], \ X^{(i)}\sim^{\text{i.i.d.}}\hat P$. For any sequence $x$ in the vocabulary $V\setminus\{O\}$, we have
\[
\hat P(X_{1: t - 1} = x_{1:t-1}) = \prod_{j=1}^{t-1}\hat P (X_j = x_j | X_{1:j-1} = x_{1:j-1})
\]
\[
= \prod_{j=1}^{t-1}\frac{ P (X_j = x_j | X_{1:j-1} = x_{1:j-1})}{ 1 - P (X_j = O | X_{1:j-1} = x_{1:j-1})}
\]
so that 
\begin{equation}
\frac{P(\hat X^{(i)}_{1:t-1})}{\hat P(\hat X^{(i)}_{1:t-1})}= \prod_{j=1}^{t-1}{( 1 - P (X_j = O | X_{1:j-1} = x_{1:j-1}))}\label{eq:3}
\end{equation}
which, when substituted into \eqref{eq:2}, yields our original expression for $\mathcal{R}$ provided in the body.

Define
\begin{equation}
    h^0_t(x) = \prod^{t-1}_{j=1}(1-P(X_j = O|X_{1:j=1}=x_{1:j-1}))
\end{equation}
\begin{equation}
    h_t(x) = P(X_t = O | X_{1:t-1}= x_{1:t-1})\mathbbm{1}_{\{T_E(x) > t-1\}}h_t^0(x)
\end{equation}
Using these definitions, $\mathcal{R}$ can be simplified as:
\begin{equation}
    \mathcal{R} = \frac{1}{n}\sum^n_{i=1}\sum_{t\geq1}h_t(\hat X^{(i)}) \label{eq:m2}
\end{equation}
Next, using our result from \eqref{eq:3}:
\begin{equation}
    \mathbb{E} [h_t(\hat X)^2] = \int h_t(\hat x_{1:t})\hat p(\hat x_{1:t})d\hat x_{1:t}\label{eq:start}
\end{equation}
\begin{equation}
    = \int h_t(\hat x_{1:t})\frac{\hat p(\hat x_{1:t})}{p( x_{1:t})} p( x_{1:t})d\hat x_{1:t}
\end{equation}
\begin{equation}
    = \int f_t(\hat x_{1:t})^2h_t^0(x_{1:t})^2\frac{1}{h_t^0(\hat x_{1:t})} p( x_{1:t})d\hat x_{1:t} = \mathbb{E}[f_t(X)^2h_t^0(X)]\label{eq:end}
\end{equation}
Since $0\leq h_s(x)\leq 1$, we can conclude that:
\begin{equation}
    \mathbb{E}[h_t(\hat X)^2] \leq \mathbb{E}[f_t(X)^2]    \label{eq:sqexp}
\end{equation}
Since $\mathcal{S}$ and $\mathcal{R}$ have the same expectation as unbiased estimators of the same probability, \eqref{eq:sqexp} demonstrates that:
\begin{equation}
    \Var(h_t(\hat X)) \leq \Var(f_t(\hat X)) 
\end{equation}
Next, for $s<t$, we have
\begin{equation}
\text{Cov}(h_s(\hat X), h_t(\hat X)) = \mathbb{E}[h_s(\hat X)h_t(\hat X)] - \mathbb{E}[h_s(\hat X)]\mathbb{E}[h_t(\hat X)]
\end{equation}
By a nearly identical argument to the one presented in \eqref{eq:start} to \eqref{eq:end}, it can be shown that

\[
    \mathbb{E}[h_s(\hat X)h_t(\hat X)] - \mathbb{E}[h_s(\hat X)]\mathbb{E}[h_t(\hat X)] = \mathbb{E}[f_s(X)f_t(X)h_s^0(X)] - \mathbb{E}[f_s(X)]\mathbb{E}[f_t(X)]
\]
\begin{equation}
    \text{Cov}(h_s(\hat X), h_t(\hat X))\leq \text{Cov}(f_s(X), f_t(X))
\end{equation}
And, by \eqref{eq:m1} and \eqref{eq:m2}, we reach our final conclusion that:
\[
    \Var(\mathcal{R})\leq \Var(\mathcal{S})
\]
\newpage
\section{ETHOS-ARES Details}
\label{app:abt_ethos}
For this empirical testing, we reproduced the entire MIMIC-IV 3.1 ETHOS-ARES pipeline to create a cohort of EHR patients, generate tokenized timelines for these patients, and predict various outcomes for these patients. In order to ensure a fair comparison between the original Monte Carlo method and our new methods, the original machine learning pipeline presented in the paper was preserved and reproduced in as many respects as possible. Notably, this includes the decision not to disable dropout during timeline inference, which is not explicitly mentioned in either \cite{renc_zero_2024} or \cite{Ren25} but is demonstrated by the fact that eval mode is not enabled for the model during inference.

However, in order to overcome the excessive expected inference times, the inference methods presented in the original ETHOS-ARES database were significantly overhauled with the aim of reducing the computational cost. First, KV caching was added to their inference methods in order to significantly reduce inference times. Additionally, since the ETHOS-ARES model uses a fixed-length sliding context window, it was necessary to add a skip size parameter, which determined how many tokens the sliding context window would jump when the timeline reached the maximum length, in order to preserve the speedups offered by KV caching for sequences that exceed the maximum context length. For this parameter, a skip size of $64$ was selected as it was the lowest parameter tried that preserved the substantial speedup offered by KV-caching.  These changes resulted in a roughly nine fold decrease in inference time. The specific run-times used to generate this estimate can be found in the log files of the code supplement and are listed in Appendix(\ref{sec.inf_time_red}). The code for these modified inference methods can be found in the GitHub repository. 

\newpage
\section{ETHOS-ARES Empirical Results}
\label{app:ea_empirical}
\begin{figure}[h!]
\centering
    {\smaller Hospital Mortality\\ \ \\}
    \hrule
    \begin{minipage}[b]{0.25\textwidth}
        \centering
        \includegraphics[width=\textwidth]{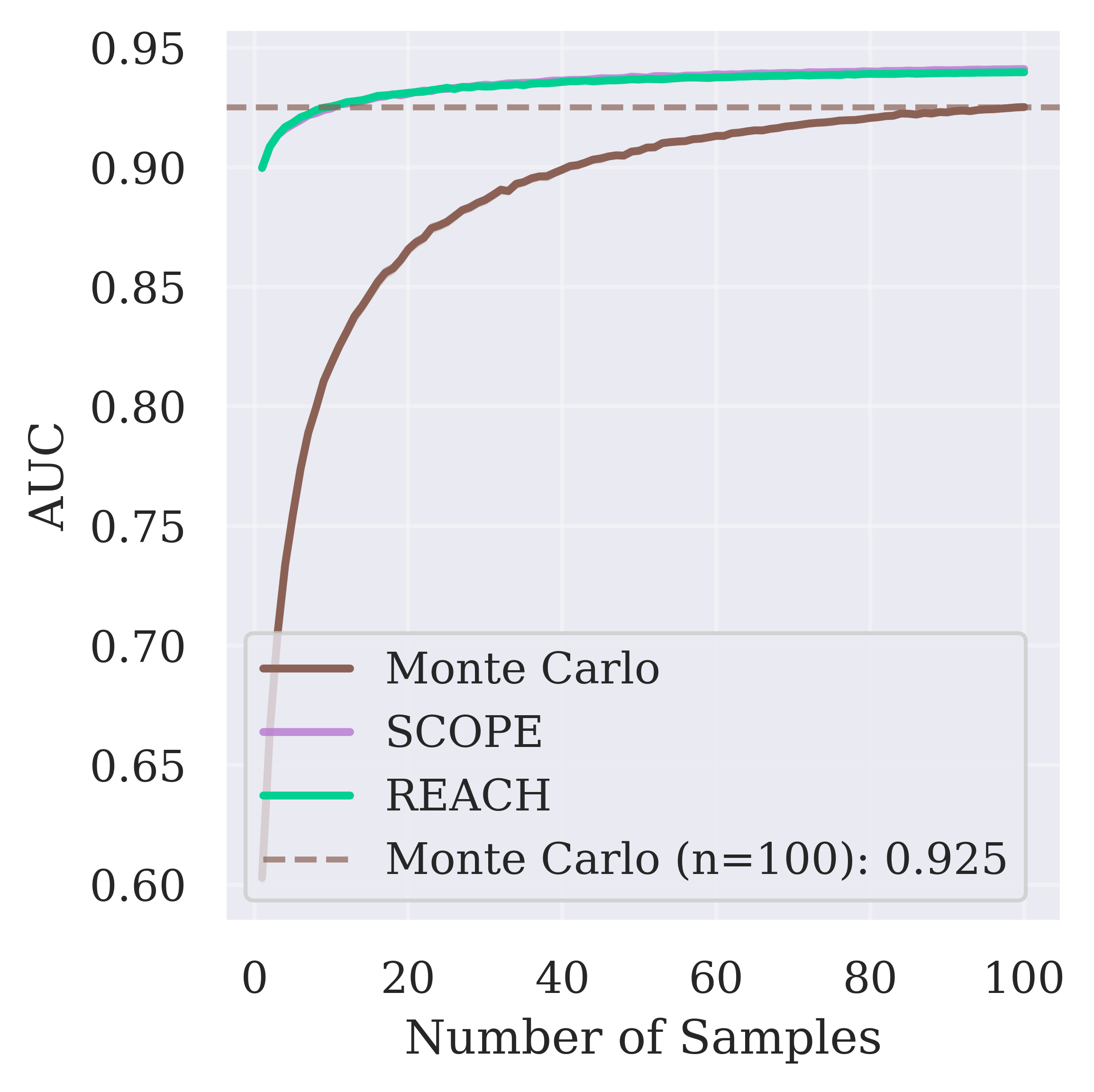}%
        \raisebox{0.45\textwidth}{\llap{\textbf{(a)}\hspace{.5em}}}%
    \end{minipage}%
    \begin{minipage}[b]{0.25\textwidth}
        \centering
        \includegraphics[width=\textwidth]{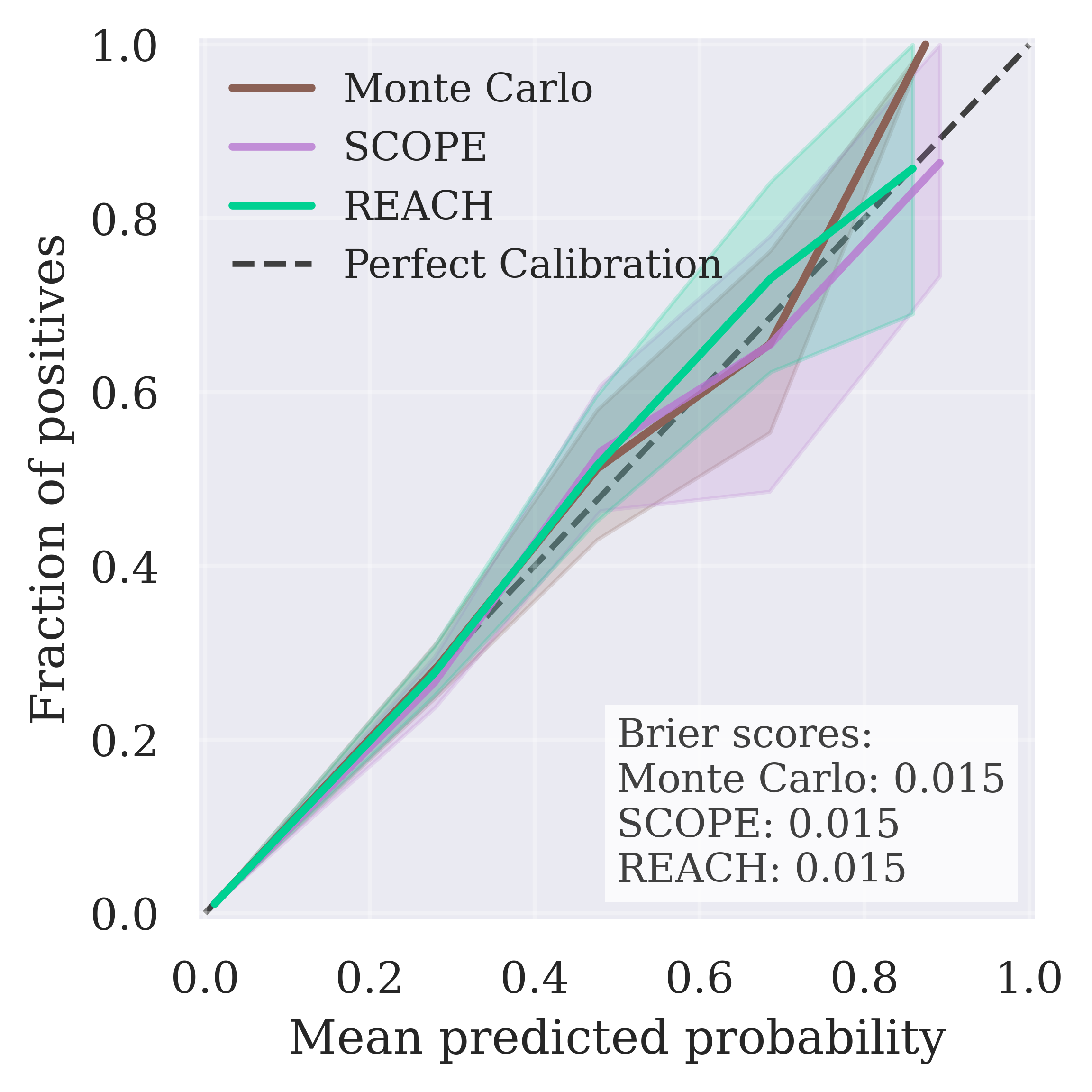}%
        \raisebox{0.45\textwidth}{\llap{\textbf{(b)}\hspace{.7em}}}%
    \end{minipage}
    \vspace{-2ex}
    \begin{minipage}[b]{0.4\textwidth}
        \centering
        \includegraphics[width=\textwidth]{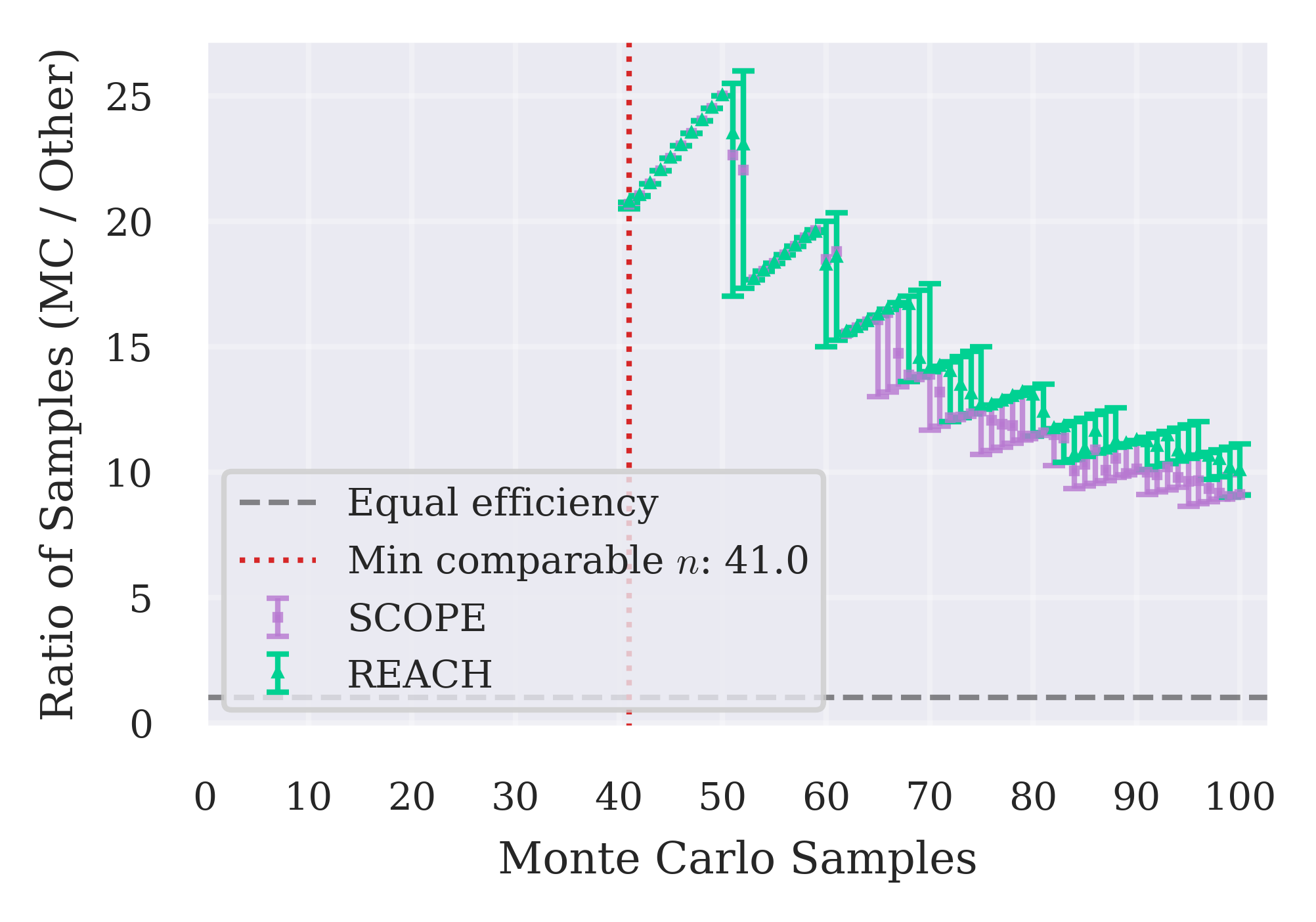}%
        \raisebox{0.45\textwidth}{\llap{\textbf{(i)}\hspace{.7em}}}%
    \end{minipage}\\ \ \\
    {\smaller ICU Admission\\ \ \\}
    \hrule
    \begin{minipage}[b]{0.25\textwidth}
        \centering
        \includegraphics[width=\textwidth]{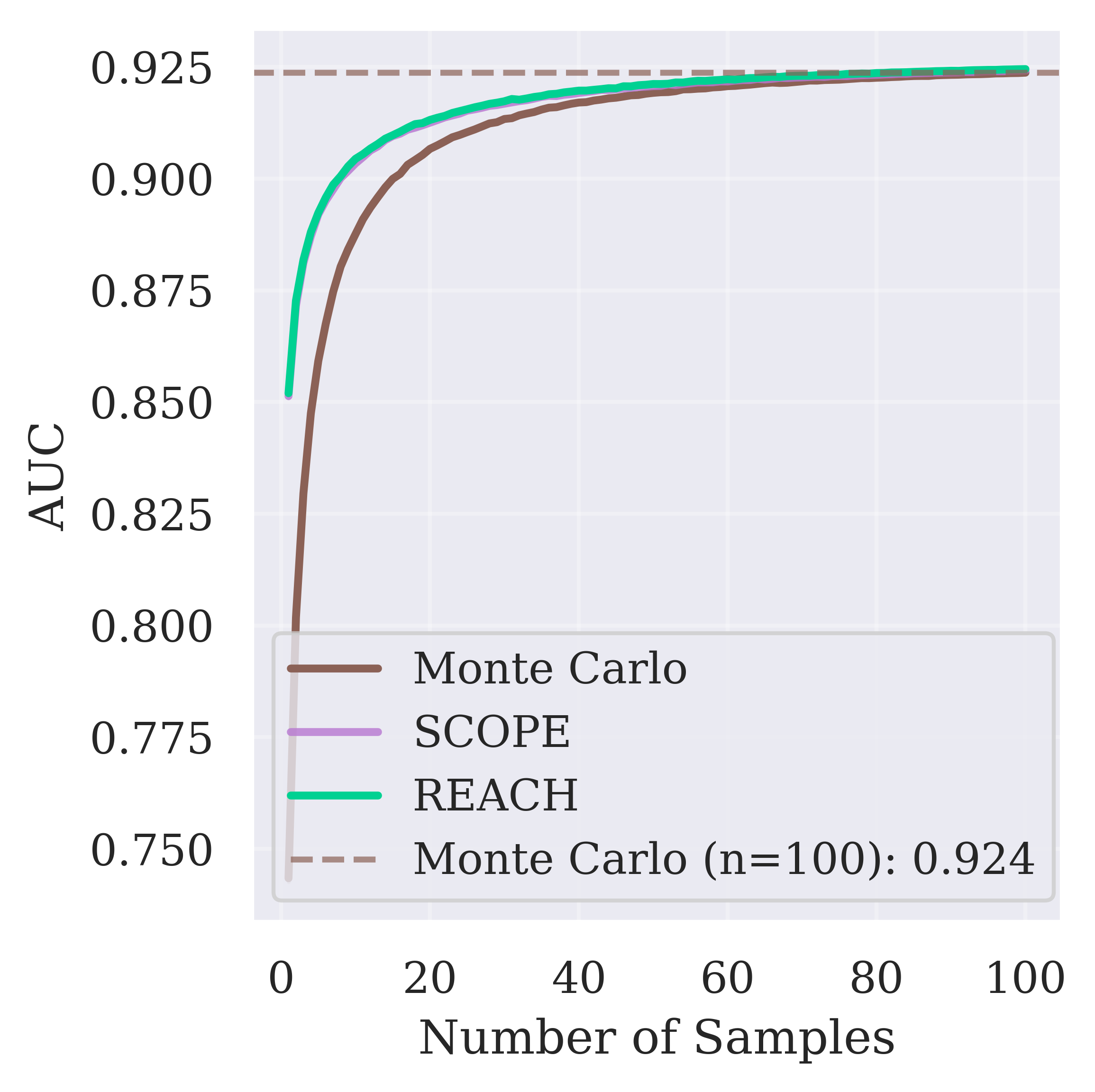}%
        \raisebox{0.45\textwidth}{\llap{\textbf{(c)}\hspace{.5em}}}%
    \end{minipage}%
    \begin{minipage}[b]{0.25\textwidth}
        \centering
        \includegraphics[width=\textwidth]{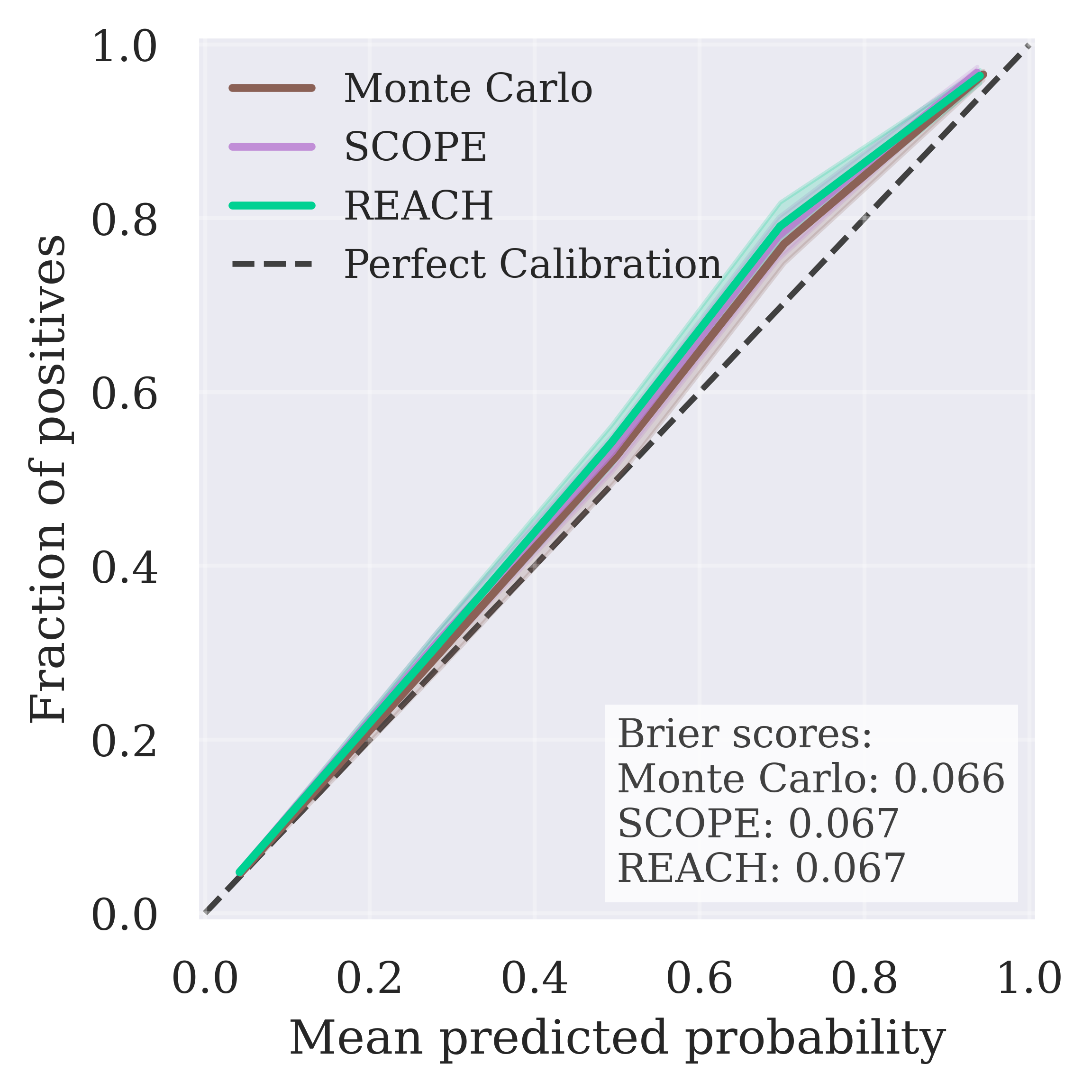}%
        \raisebox{0.45\textwidth}{\llap{\textbf{(d)}\hspace{.7em}}}%
    \end{minipage}
    \vspace{-2ex}
    \begin{minipage}[b]{0.4\textwidth}
        \centering
        \includegraphics[width=\textwidth]{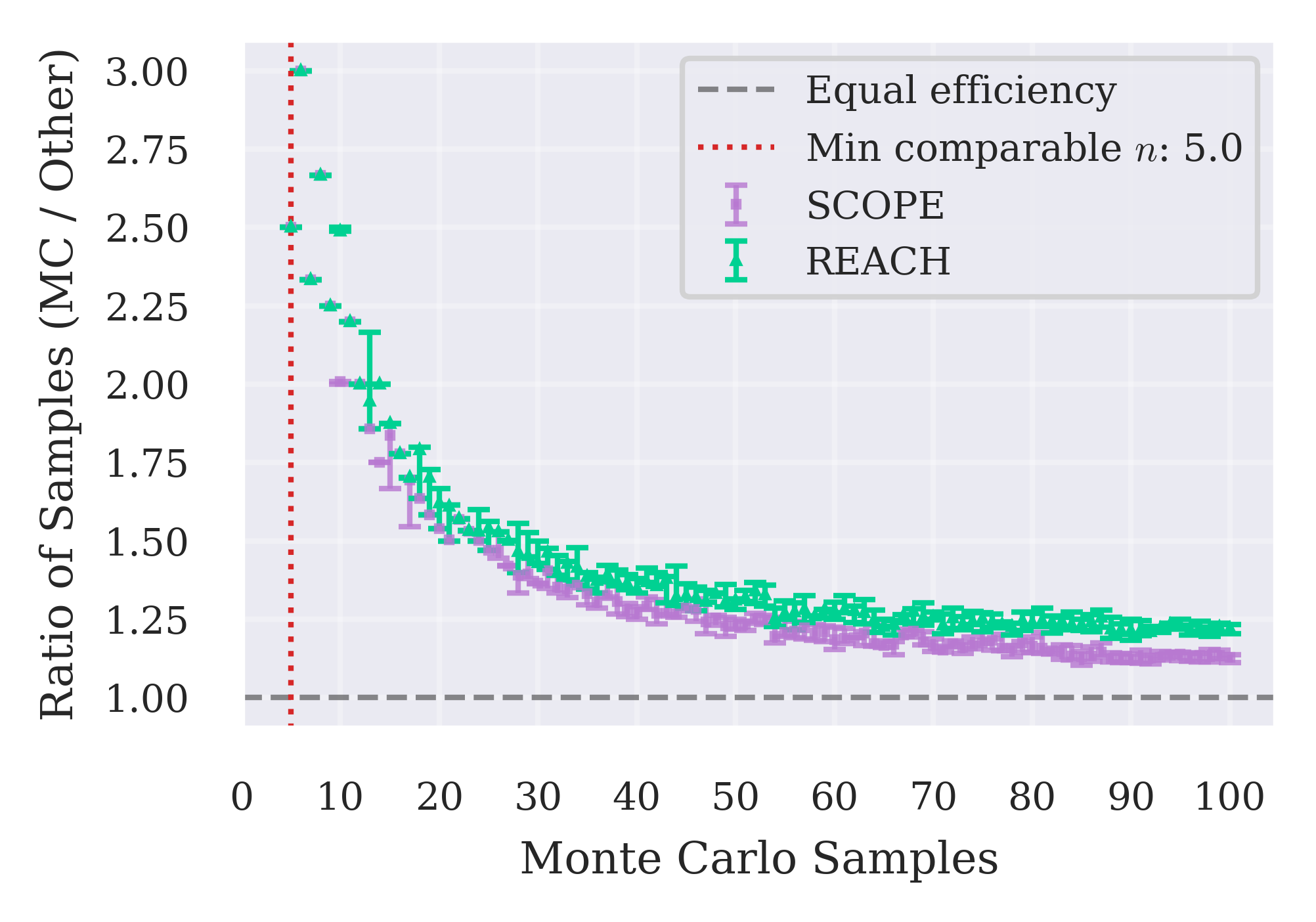}%
        \raisebox{0.45\textwidth}{\llap{\textbf{(ii)}\hspace{.7em}}}%
    \end{minipage}\\
    \caption{\footnotesize(a) Hospital Mortality AUC vs number of sampled timelines. (b) Hospital Mortality calibration curves (100 samples for all estimators). (c) ICU Admission AUC vs number of sampled timelines. (d) ICU Admission calibration curves (100 samples for all estimators).}
    \label{fig:ETHOS-fig}
\end{figure}

Beyond the abstract understanding of these estimators offered by our proofs and Markov chain experiments, the ETHOS-ARES experiments examine the effectiveness of these novel estimators on real world tasks and data. Figures~\ref{fig:ETHOS-fig}(a) and ~\ref{fig:ETHOS-fig}(i) show that, using only about $11$ ($95\% $ CI: $[11, 11]$) predicted future timelines for SCOPE and $10$ ($95\% $ CI: $[9, 11]$) for REACH, both methods matched the performance of the Monte Carlo estimator that utilized $100$ predicted timelines for the task of predicting hospital mortality. This represents a drastic reduction in the amount of computational resources required to compute such estimates. Furthermore, this reduction in the required inference for equivalent performance becomes even more dramatic with lower sample counts. With $40$ or fewer sampled timelines, Monte Carlo fails to match the performance of either SCOPE or REACH using only $1$ sampled timeline.  

Although much less dramatic than in the hospital mortality task, ~\ref{fig:ETHOS-fig}(c) and ~\ref{fig:ETHOS-fig}(ii) still show both SCOPE and REACH offer improvements in predictive accuracy over the Monte Carlo estimator at all sample counts when predicting ICU admission, with SCOPE matching $100$ sample performance with only $89$ ($95\% $ CI: $[88, 90]$) and REACH achieving the same in only $82$ ($95\% $ CI: $[81, 83]$). Just as in the case of hospital mortality, the ratio of samples increases with smaller reference sample counts, with Monte Carlo requiring $5$ samples to best the performance of both estimators with only $1$ timeline.

For both of the above tasks,  ~\ref{fig:ETHOS-fig}(b) and  ~\ref{fig:ETHOS-fig}(d) demonstrate that there is no significant degradation in the quality of the calibration in either predictive task. Additionally, for all combinations of task and estimator except for SCOPE on hospital mortality with $11$ samples, there were no statistically significant differences in Brier scores of the Monte Carlo method using $100$ samples compared to the equivalent sample counts presented above, and, although statistically significantly worse, the Brier score of the $11$-sample SCOPE estimate on hospital mortality is not meaningfully different from the Monte Carlo $100$-sample score  (0.016 vs 0.015).

The discrepancy between the impact of SCOPE and REACH in these two tasks can, in part, be understood as a result of the ``spontaneity" of each task, reflected in Figure~\ref{fig:markov-experiments}(c). Depending on the tokenization process, mortality can be a rather sudden event in a patient timeline, and it will therefore continuously carry some ambient probability of being the next token based on the risk suggested by the past timeline. On the other hand, clinician initiated actions such as ICU admission are precipitated by clinical information that should be present in the tokenized timeline. In fact, the empirical spontaneity of ICU admission was $0.00583$ while it was $0.00729$ for hospital mortality, an over $25\%$ increase. So, by~\eqref{eq:spont}, this would suggest that the improvement over the Monte Carlo method would be less significant, since the outcome free tokenized timeline would be much less uncertain. For future work, this suggests that the greatest benefit of these methods will be found on tasks that involve a greater degree of spontaneity. In addition to spontaneity, ICU admission is a much more prevalent outcome than mortality ($15.44\%$ vs $1.85\%$, \citep{Ren25}). This means that resolving the dispersion of estimates problem is much less meaningful than it is for a rarer event, like hospital mortality. 
\newpage
\section{ETHOS-ARES: Total Inference Time Reductions}
\label{sec.inf_time_red}
In order to compute the amount of GPU time required, the total run times of our 32 data splits for the hospital mortality task using the M2 estimator were summed. 
\begin{table}[h]
\centering
\caption{Runtime Summary for ethos\_M2\_ed\_4334712 (Files 0-31)}
\label{tab:runtime}
\begin{tabular}{|c|c|}
\hline
\textbf{File Index} & \textbf{Runtime (H:MM:SS)} \\
\hline
0 & 6:54:45 \\
1 & 6:32:39 \\
2 & 6:49:25 \\
3 & 7:31:00 \\
4 & 6:40:11 \\
5 & 6:48:31 \\
6 & 6:56:42 \\
7 & 7:39:13 \\
8 & 6:44:45 \\
9 & 5:57:16 \\
10 & 8:22:20 \\
11 & 6:04:10 \\
12 & 6:59:03 \\
13 & 7:31:58 \\
14 & 6:44:37 \\
15 & 7:57:52 \\
16 & 7:54:19 \\
17 & 7:22:43 \\
18 & 7:03:45 \\
19 & 7:39:57 \\
20 & 6:35:22 \\
21 & 8:16:21 \\
22 & 7:26:34 \\
23 & 7:09:57 \\
24 & 7:43:19 \\
25 & 7:45:51 \\
26 & 6:52:54 \\
27 & 7:01:50 \\
28 & 7:54:10 \\
29 & 6:31:27 \\
30 & 6:39:40 \\
31 & 6:54:19 \\
\hline
\textbf{Total} & \textbf{164:33:28} \\
\hline
\end{tabular}
\end{table}

Next, in order to estimate the time required by the standard ETHOS-ARES library, the inference code was left to run with 8xA100 GPUS for 1 hour. This generated $22395$ samples out of the required $4304700$. This implies a little over $192$ hours will be required to complete inference using this setup. Multiplying by $8$ to determine the GPU hours required, we get over $\fbox{1537}$ GPU hours. This yields a total speedup factor:

$$\frac{1537.74}{164.55} = \fbox{9.35}$$

Note that this does not include any speedups from the reductions in the number of samples required for equivalent performance between methods. For the sake of transparency, the log files that contain these measurements are made available in the codebase.

\clearpage
\section{ETHOS-ARES Hospital Mortality: Performance vs Sample Count Plots and P-Values}
\label{app:mortality_perf}

\begin{figure}[!htb]
\centering
    Hospital Mortality
    \includegraphics[width=\textwidth]{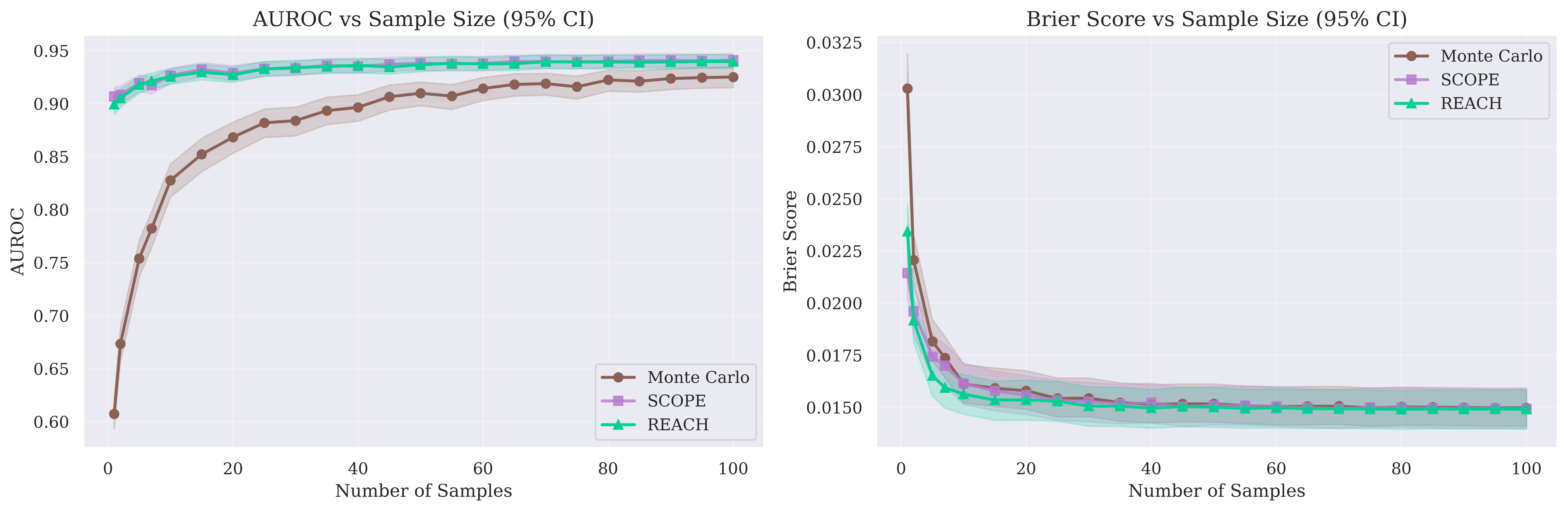}
\end{figure}
\begin{table}[htbp]
\centering
\caption{Hospital Mortality prediction performance comparison}
\label{tab:hospital_mortality}
\begin{tabular}{lccc}
\toprule
Method & Number of Samples & ROC-AUC & Brier Score \\
\midrule
Monte Carlo & 100 & 0.925 [0.915, 0.935] & 0.015 [0.014, 0.016] \\
SCOPE       & 11  & 0.926 [0.919, 0.934] & 0.016 [0.015, 0.017] \\
REACH       & 10  & 0.925 [0.917, 0.932] & 0.016 [0.015, 0.017] \\
\midrule
\multicolumn{4}{l}{\textit{One-sided permutation test p-values}} \\
\midrule
MC vs SCOPE & - & 0.550 & 0.030 \\
MC vs REACH & - & 0.477 & 0.135 \\
\bottomrule
\end{tabular}
\end{table}
\begin{figure}[!htb]
\centering
    Calibration curves with $n=11$ for SCOPE and $n=10$ for REACH
    \includegraphics[width=0.4\textwidth]{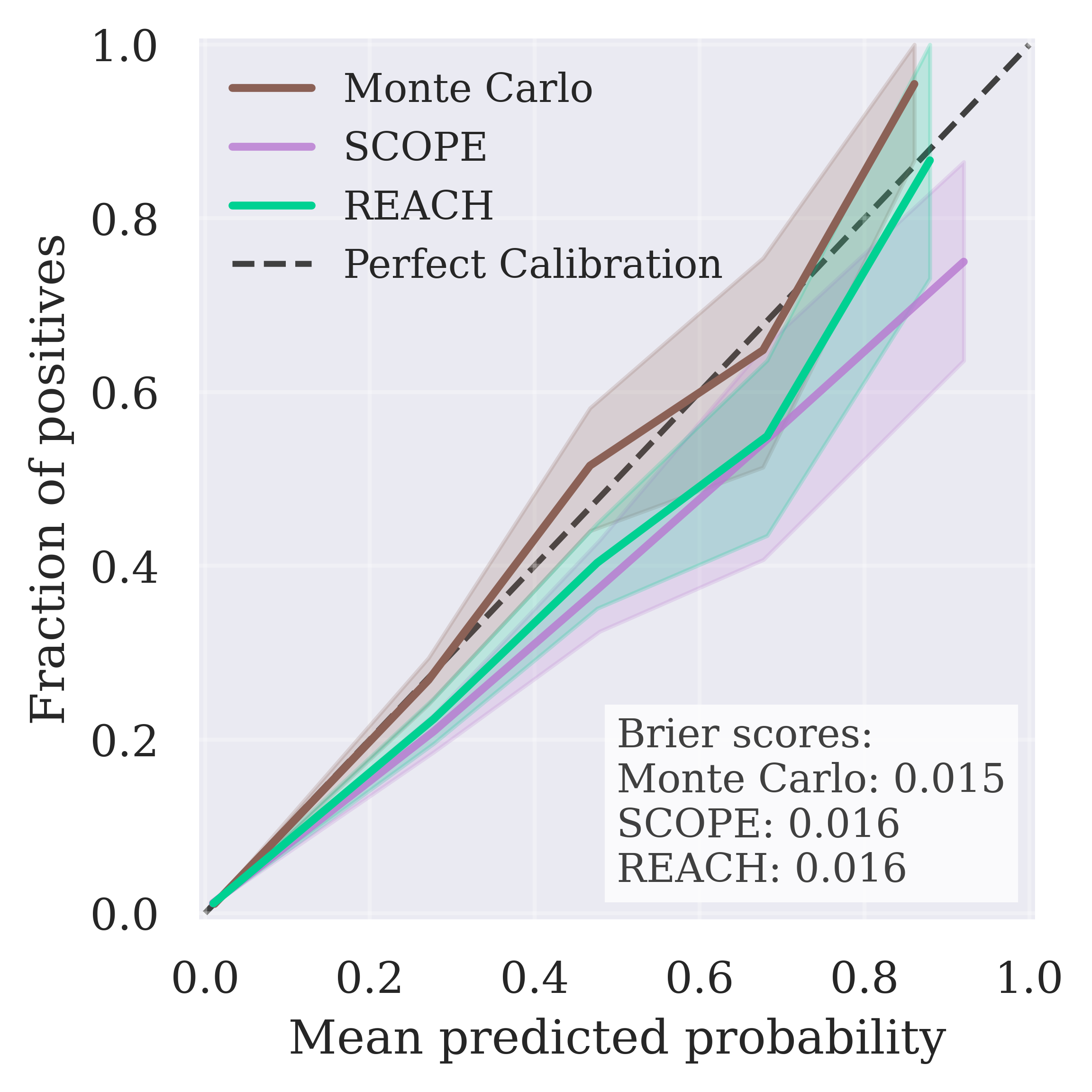}
\end{figure}

\clearpage
\section{ETHOS-ARES ICU Admission: Performance vs Sample Count Plots and P-Values}
\label{app:icu_adm_perf}
\begin{figure}[!htb]
    \centering
    ICU Admission
    \includegraphics[width=\textwidth]{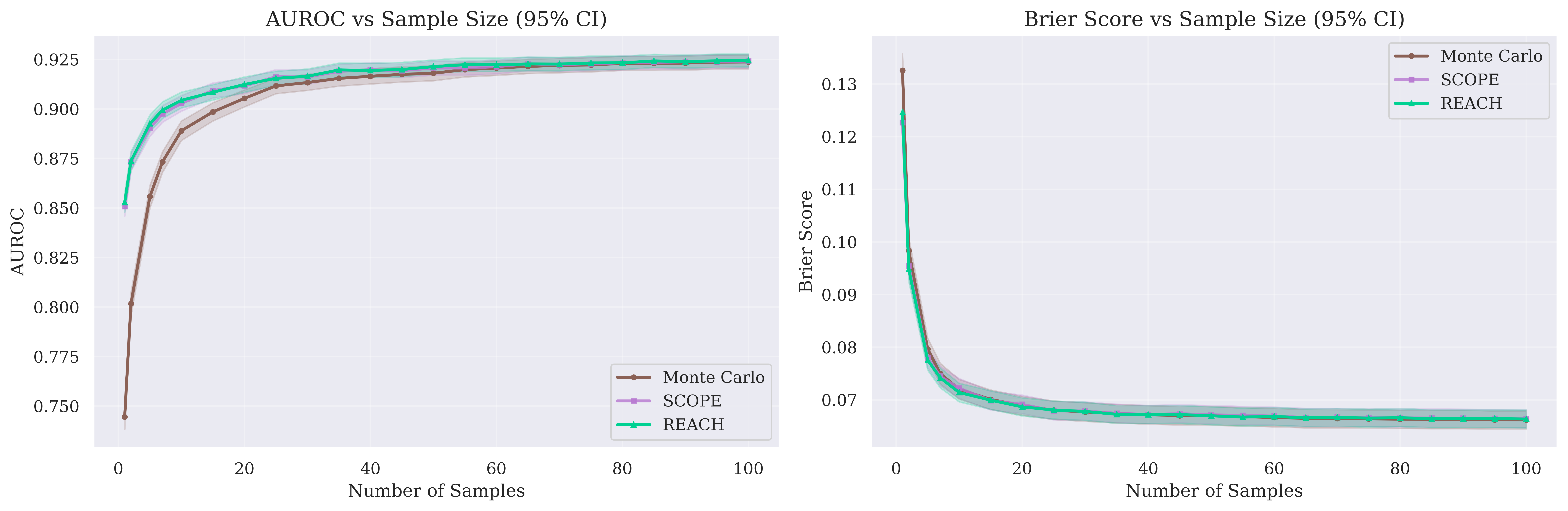}
\end{figure}

\begin{table}[htbp]
\centering
\caption{ETHOS-ARES: ICU Admission prediction performance comparison}
\label{tab:icu_admission}
\begin{tabular}{lccc}
\toprule
Method & Number of Samples & ROC-AUC & Brier Score \\
\midrule
Monte Carlo & 100 & 0.924 [0.920, 0.927] & 0.066 [0.065, 0.068] \\
SCOPE & 89 & 0.923 [0.920, 0.927] & 0.066 [0.065, 0.068] \\
REACH & 82 & 0.923 [0.920, 0.927] & 0.067 [0.065, 0.068] \\
\midrule
\multicolumn{4}{l}{\textit{One-sided permutation test p-values}} \\
\midrule
MC vs SCOPE & - & 0.453 & 0.411 \\
MC vs REACH & - & 0.407 & 0.404 \\
\bottomrule
\end{tabular}
\end{table}

\begin{figure}[!htb]
\centering
    Calibration curves with $n=89$ for SCOPE and $n=82$ for REACH
    \includegraphics[width=0.4\textwidth]{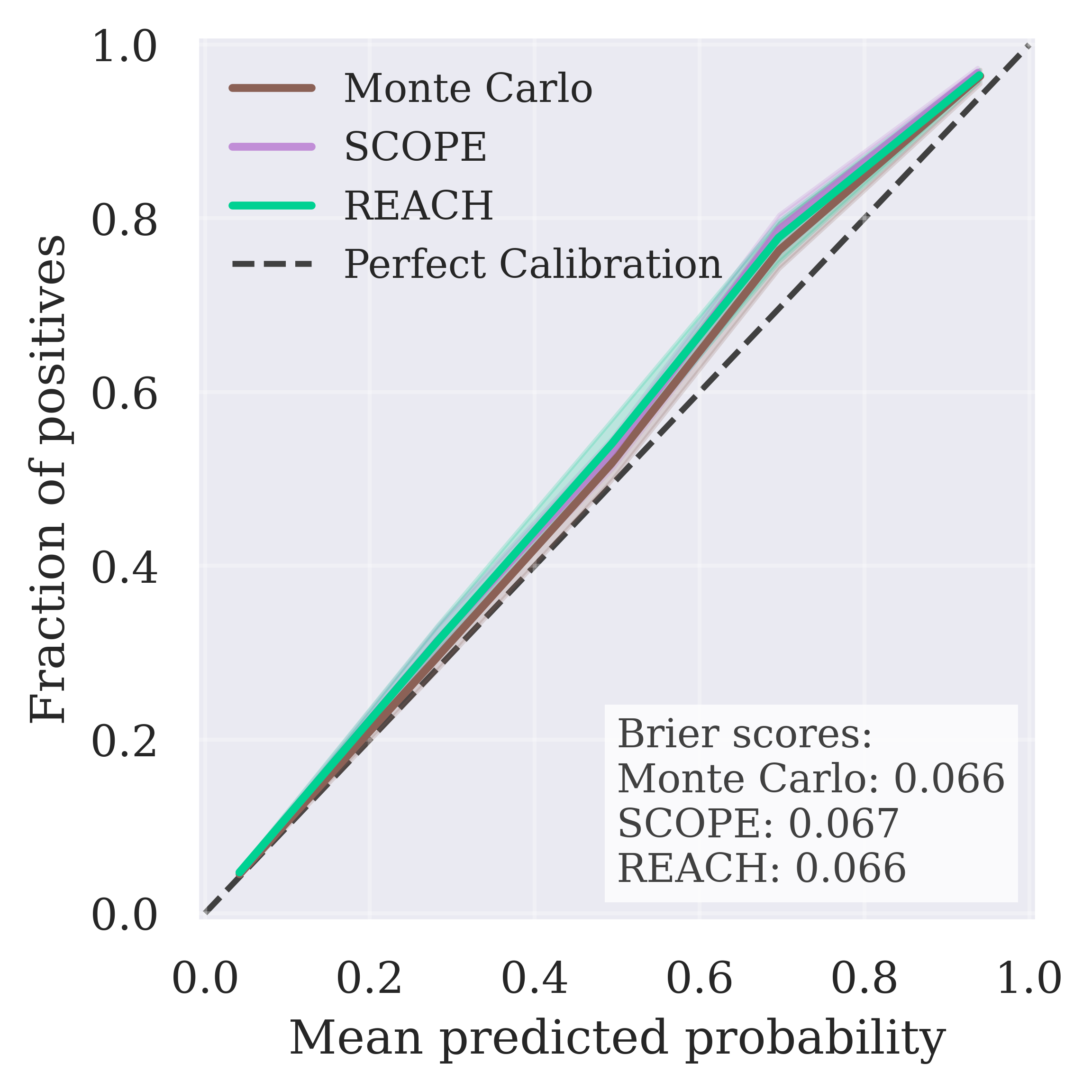}
\end{figure}

\newpage
\section{Counterexample demonstrating the impossibility of a positive bound \texorpdfstring{$M$}{M} such that if 
\label{app:counter}
\texorpdfstring{$P(A) <M$}{P(A) < M} then \texorpdfstring{$\Var(M_0) > \Var(\mathcal{S})$}{Var(M0) > Var(S)} for all models and stopping conditions}
In the body, it was demonstrated that $\Var(\mathcal{S})$ is not necessarily less than $\Var(M_0)$. Specifically, timelines for which the $\mathcal{S}$ sub-estimator predicts an estimate for the value of the probability above $1$ contribute to the inequality $\Var(M_0) < \Var(\mathcal{S})$. Given this fact, it seemed plausible that there could exist some positive lower bound $M$ for the probability of the outcome of interest, $P(A)$, for which the following could hold: for any model and stopping condition, if $P(A) < M$ then $\Var(\mathcal{S}) < \Var(M_0)$. Such a bound would be quite useful as it would allow you determine that $\mathcal{S}$ would be a lower variance estimator than $M_0$ based on the probability of the outcome alone.

However, this fact is not true and is shown to be false by the following counterexample. 

Consider the following model and stopping condition. At first, the model has two possible token outputs $A$ and $B$. Let it generate token $A$ with arbitrary probability $p$, and $B$ otherwise. Neither contribute toward the time limit or are the outcome of interest. The model then operates according to the following rules:
\begin{itemize}
    \item If $A$ was generated at the first step:
        \item Immediately terminate the timeline without generating any further tokens.
    \item If $B$ was generated at the first step:
        \item Generate new tokens with equal probability of it being token $H$ or token $T$. Terminate either after $3$ tokens are generated or after the appearance of the first heads. 
\end{itemize}

For this counterexample, we will treat $H$ as the outcome of interest token. Naturally:
\begin{equation}
    P(H \in X) = \frac{7}{8}(1-p)
\end{equation}

Since $\mathcal{S}$ and $M_0$ are unbiased estimators for this probability, this likewise is the expectation of both of these random variables. In order to compute the difference in variance between $\mathcal{S}$ and $M_0$, we can proceed directly from the definition of variance and the unbiasedness of our estimators:
\begin{equation}
    \Var(\mathcal{S}) - \Var(M_0) = \mathbb{E}(\mathcal{S}^2) - \mathbb{E}(M_0^2)
\end{equation}
\begin{table}[htpb]
    \centering
    \begin{tabular}{c|c|c|c}
        Outcome & Probability & $M_0^2$ contribution & $\mathcal{S}^2$ contribution\\
        \hline
         A & $p$& 0 & 0\\
        BH & $\frac{1}{2}(1-p)$ & $1$ & $\frac{1}{4}$ \\
        BTH & $\frac{1}{4}(1-p)$ & $1$ & $1$ \\
        BTTH & $\frac{1}{8}(1-p)$ & $1$ & $\frac{9}{4} $ \\
        BTTT & $\frac{1}{8}(1-p)$ & $0$ & $\frac{9}{4} $
    \end{tabular}
\end{table}
From the table:
$$\mathbb{E}(M_0^2) = \frac{7}{8}(1-p)$$
$$\mathbb{E}(\mathcal{S}^2) = \frac{15}{16}(1-p)$$
\begin{equation}
    \Var(\mathcal{S}) - \Var(M_0) = \frac{1}{16}(1-p) \label{eq:app3final}
\end{equation}  

Note that if $p<1$, \eqref{eq:app3final} implies $\Var(\mathcal{S}) - \Var(M_0) >0$. Further, if we allow $p$ get arbitrarily close to $1$, $P(H\in X) \rightarrow 0$, while the variance of $\mathcal{S}$ remains strictly worse than the variance of $M_0$. This shows that there cannot exist a positive value $M$ that guarantees that if $P(H\in X) < M$ then $\Var(\mathcal{S}) < \Var(M_0)$.

\newpage
\FloatBarrier
\section{Experimental Methods}
\label{app:mc_methods}
\subsection{Markov Chain Experiments}
In order to measure how the theoretical guarantees of these techniques translate to actual improvements in the accuracy of generative event model derived predictors, we first implement a suite of Markov Chain synthetic experiments that allow us to analyze the properties of these new estimators in an easily controlled and computationally cheap manner. Under this framework, the next token distribution relies only on the most recent token, rather than the entire sequence of tokens. For these experiments, we reduce the generative event model timeline simulation paradigm of ``Generate next tokens until a certain time limit or terminal condition'' to ``Progress in the discrete time Markov chain until a specified number of steps have been taken.'' And, standing in for a clinically significant token, we choose one state in the chain to act as the ``outcome of interest'' and use our two new methods and the traditional Monte Carlo method to estimate the probability of reaching this ``outcome of interest'' state. 

\newpage
\section{Markov Chain Synthetic Experiments}
\label{app:mc_synthetic}

\begin{figure}[h!]
    \centering
    \subfloat[\label{fig:res-dist}]{%
        \includegraphics[width=0.25\textwidth]{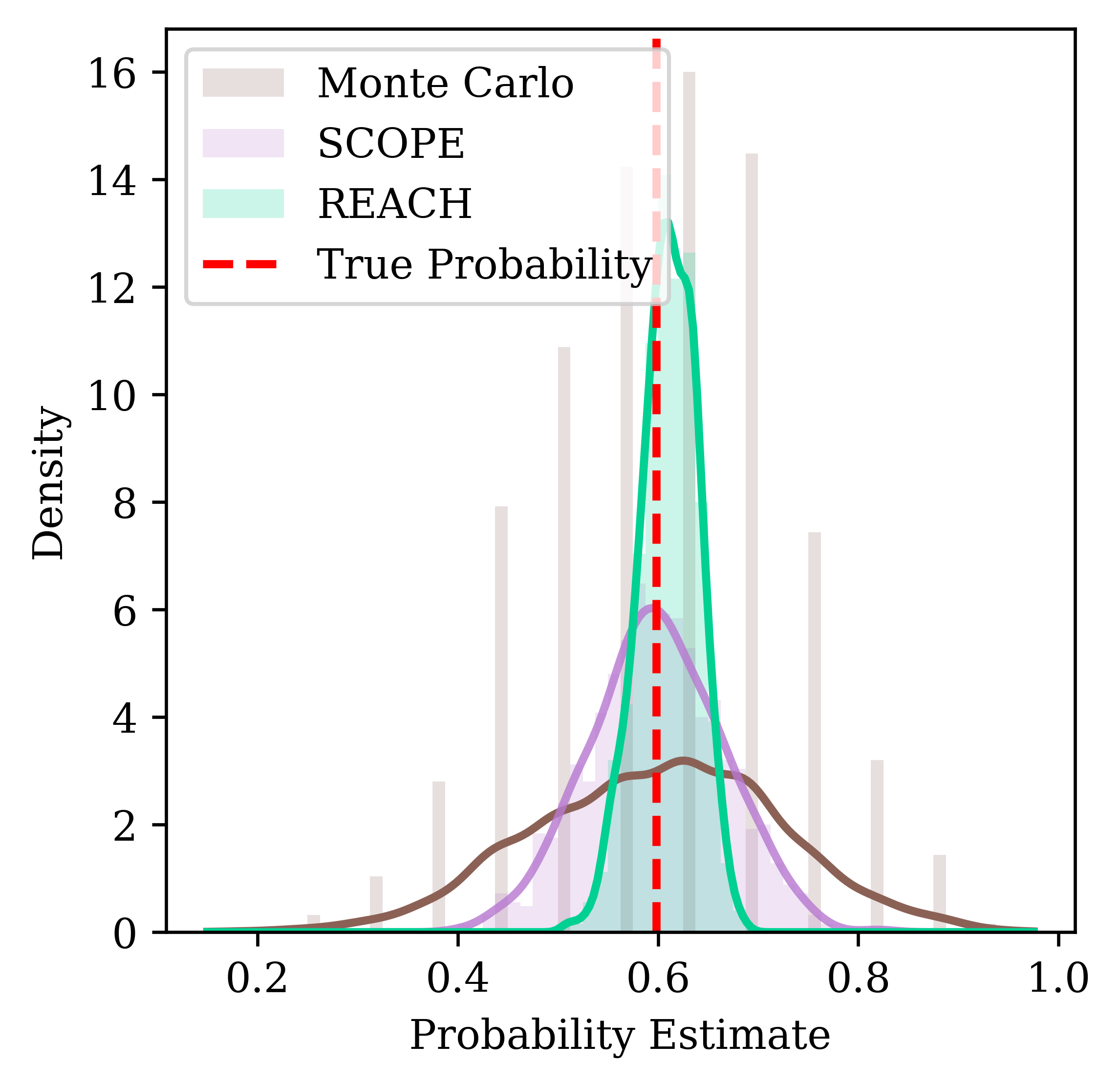}%
    }%
    \hfill
    \subfloat[\label{fig:log-probs}]{%
        \includegraphics[width=0.25\textwidth]{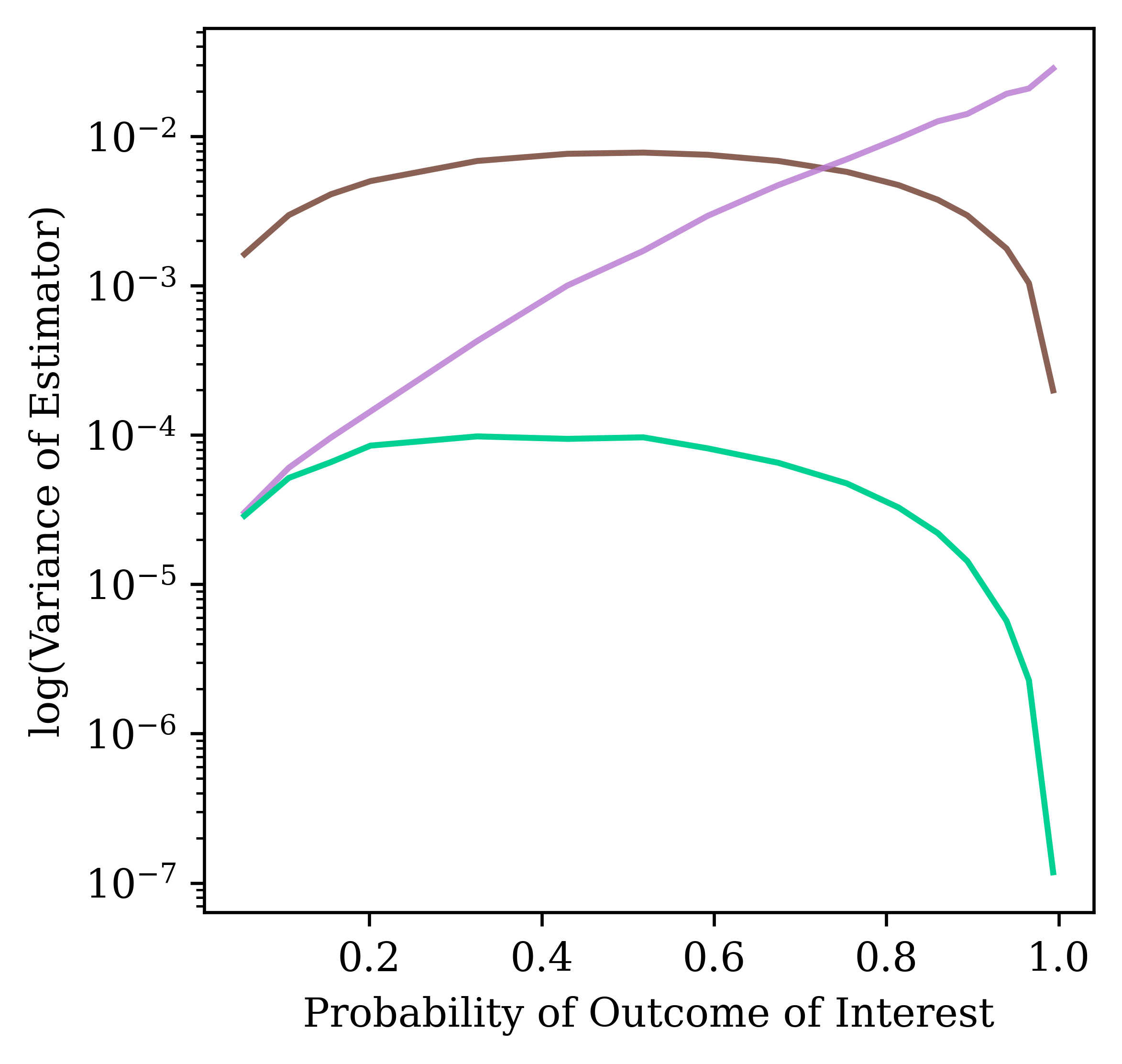}%
    }%
    \hfill
    \subfloat[\label{fig:spont}]{%
        \includegraphics[width=0.25\textwidth]{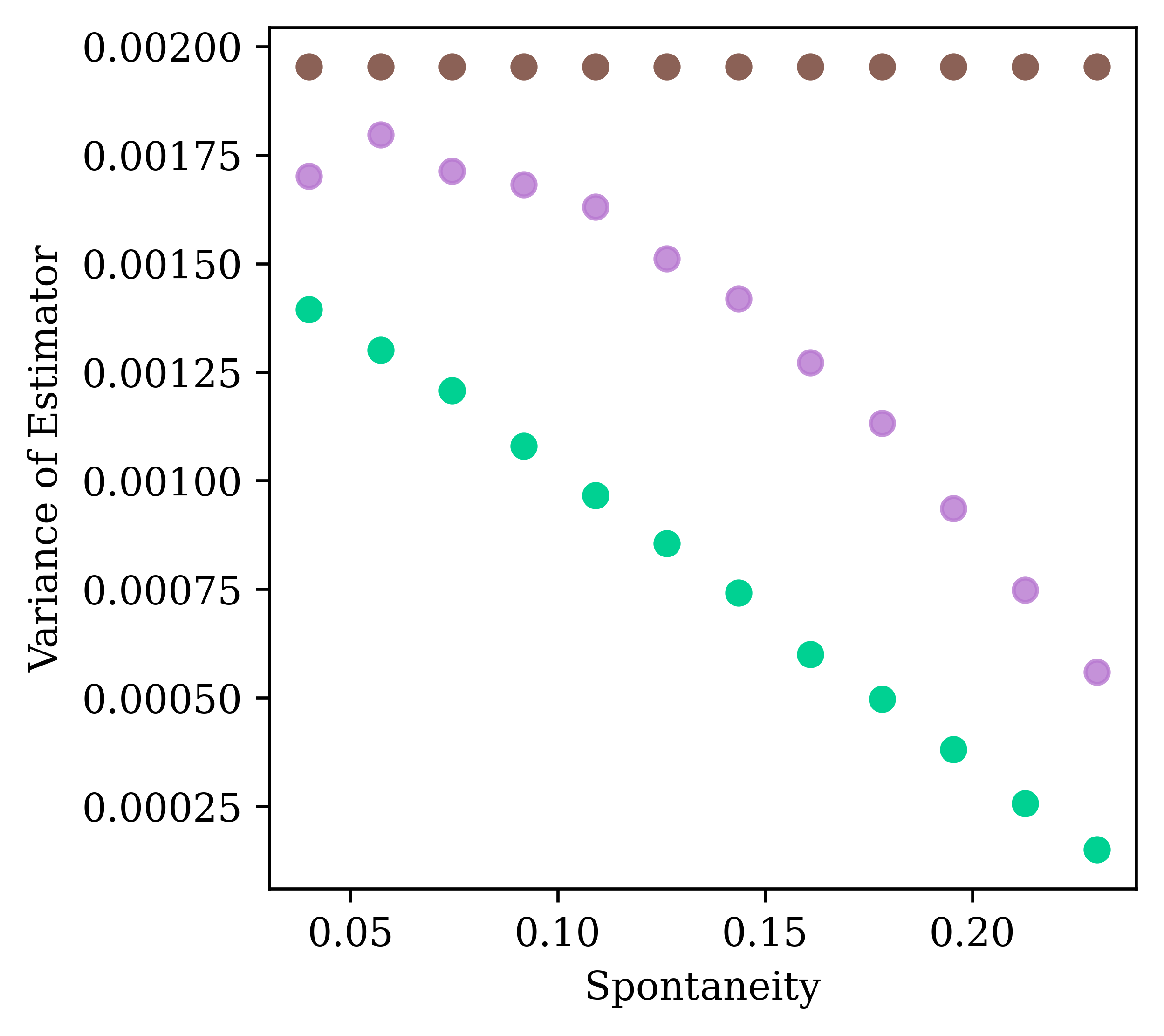}%
    }%
    \hfill
    \subfloat[\label{fig:log-samps}]{%
        \includegraphics[width=0.25\textwidth]{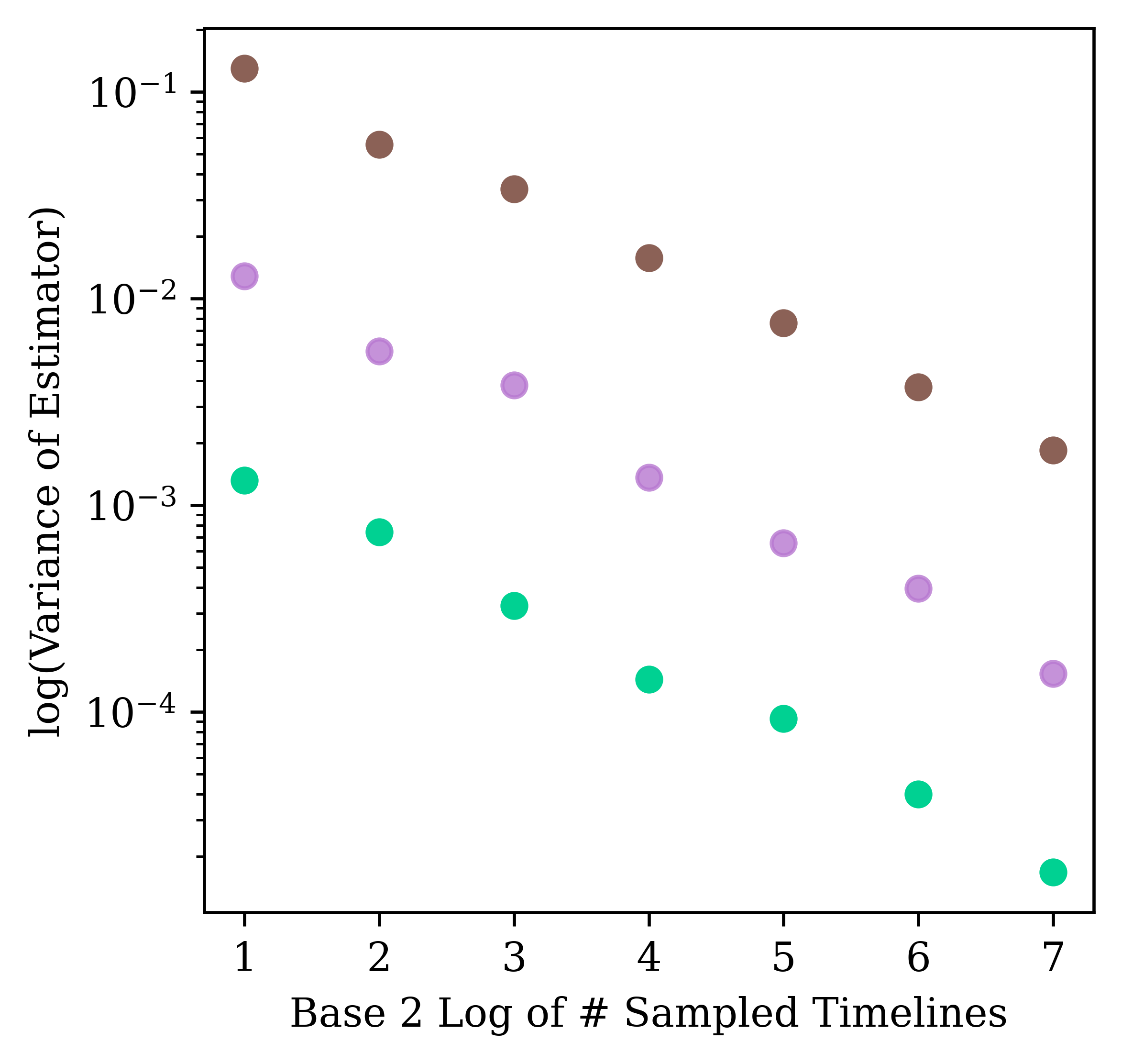}%
    }%
    \caption{%
        \textbf{(a)} Distribution of estimators vs.\ true probability.
        \textbf{(b)} Variance of estimator vs.\ probability.
        \textbf{(c)} Variance vs.\ spontaneity.
        \textbf{(d)} Variance vs.\ number of sampled timelines.
    }
    \label{fig:markov-experiments}
\end{figure}

Having rigorously demonstrated important facts about SCOPE and REACH, we now use Markov chains to demonstrate the impact of these estimators under directly controlled model and task configurations. Note: All figures presented in this section were generated using the default parameters of $4$ experiments in the Markov chain notebook.

Figure~\ref{fig:markov-experiments}(a) visually demonstrates the way in which both SCOPE and REACH resolve the dispersion of estimates problem, which can be seen in the fully separated spikes of the Monte Carlo estimator's distribution. REACH and SCOPE also visibly reduce variance over the Monte Carlo estimator.

In Figure~\ref{fig:markov-experiments}(b), the relationship between the variance of SCOPE and REACH and the probability of the outcome of interest is clearly shown. For lower true probabilities, the variance of SCOPE is extremely low and is not visibly  different from that of REACH. However, as the true probability grows, and with it the probability of $\mathcal{S}$ producing sampled values that exceed $1$, the variance of SCOPE grows rapidly and exceeds that of the MC estimator when the probability of the outcome of interest is above $0.8$.

Next, Figure~\ref{fig:markov-experiments}(c) reflects how the difference in variance between REACH and the MC estimator depends on the uncertainty of the non-outcome timelines. The Markov chains for this figure are uniquely determined by fixing both the probability of reaching the outcome of interest at $0.5$ and the spontaneity of the chain at various values between 0 and 0.25. Since spontaneity is equal to the difference in variance between the Monte Carlo and REACH methods, the linear decrease in variance of the REACH estimator exactly matches this theoretical result. However, although not directly implied by our earlier work, a similar relationship seems to hold between SCOPE and the MC estimator. Finally, Figure~\ref{fig:markov-experiments}(d), which is closely reflected experimentally in Figure~\ref{fig:ETHOS-fig}(a), examines the performance of all three methods under varying numbers of samples. Note that the $x$ axis is in log base $2$ scaling. 

\clearpage

\section{MIMIC-IV Calibration Results}
\label{app:curves_and_cal_1}
\begin{figure}[h!]
    \centering
    \includegraphics[width=\linewidth]{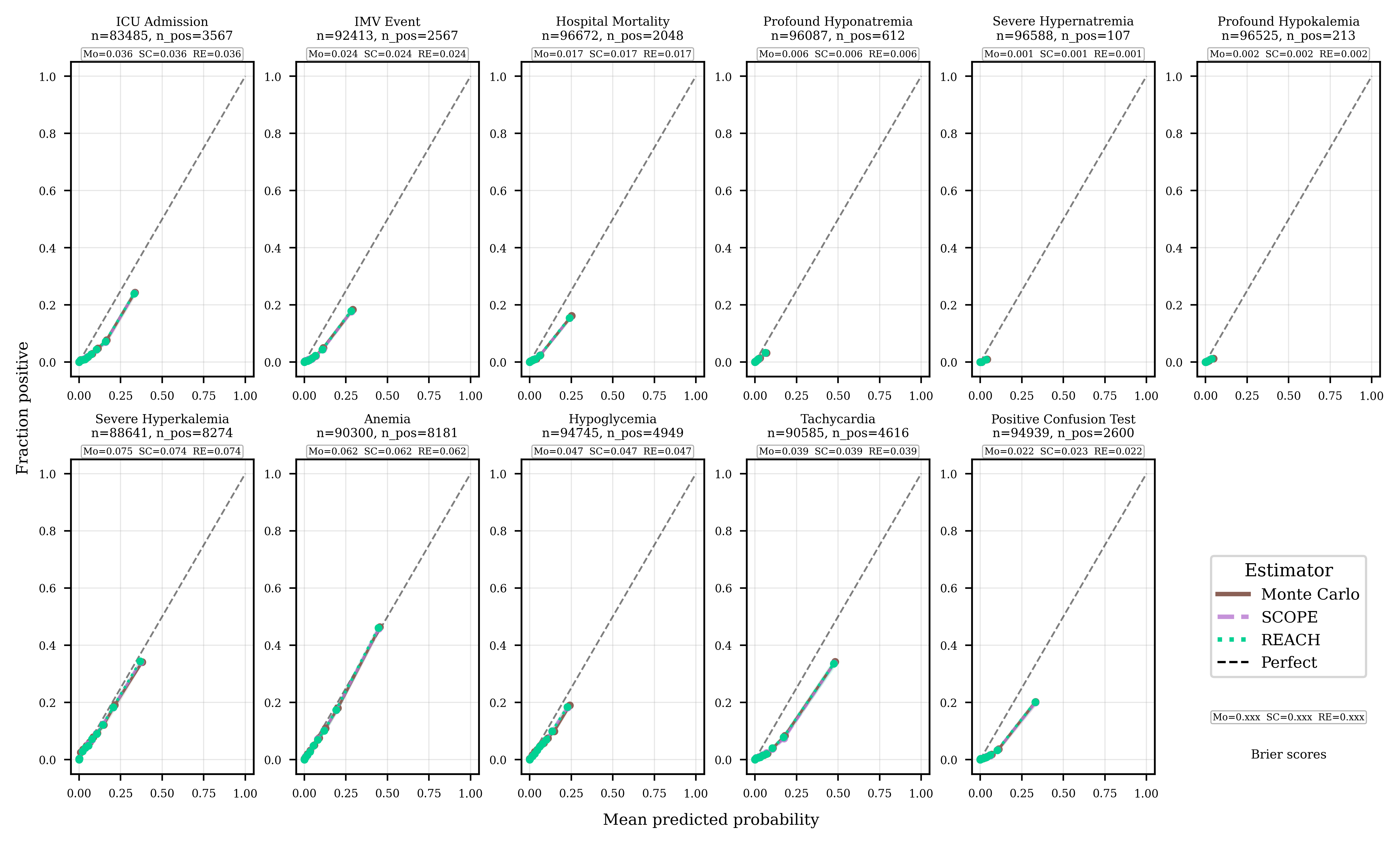}
    \caption{Brier scores and calibration curves of MIMIC-IV prediction tasks}
    \label{fig:placeholder}
\end{figure}

\section{UCMC Calibration Results}
\label{app:curves_and_cal_2}
\begin{figure}[h!]
    \centering
    \includegraphics[width=\linewidth]{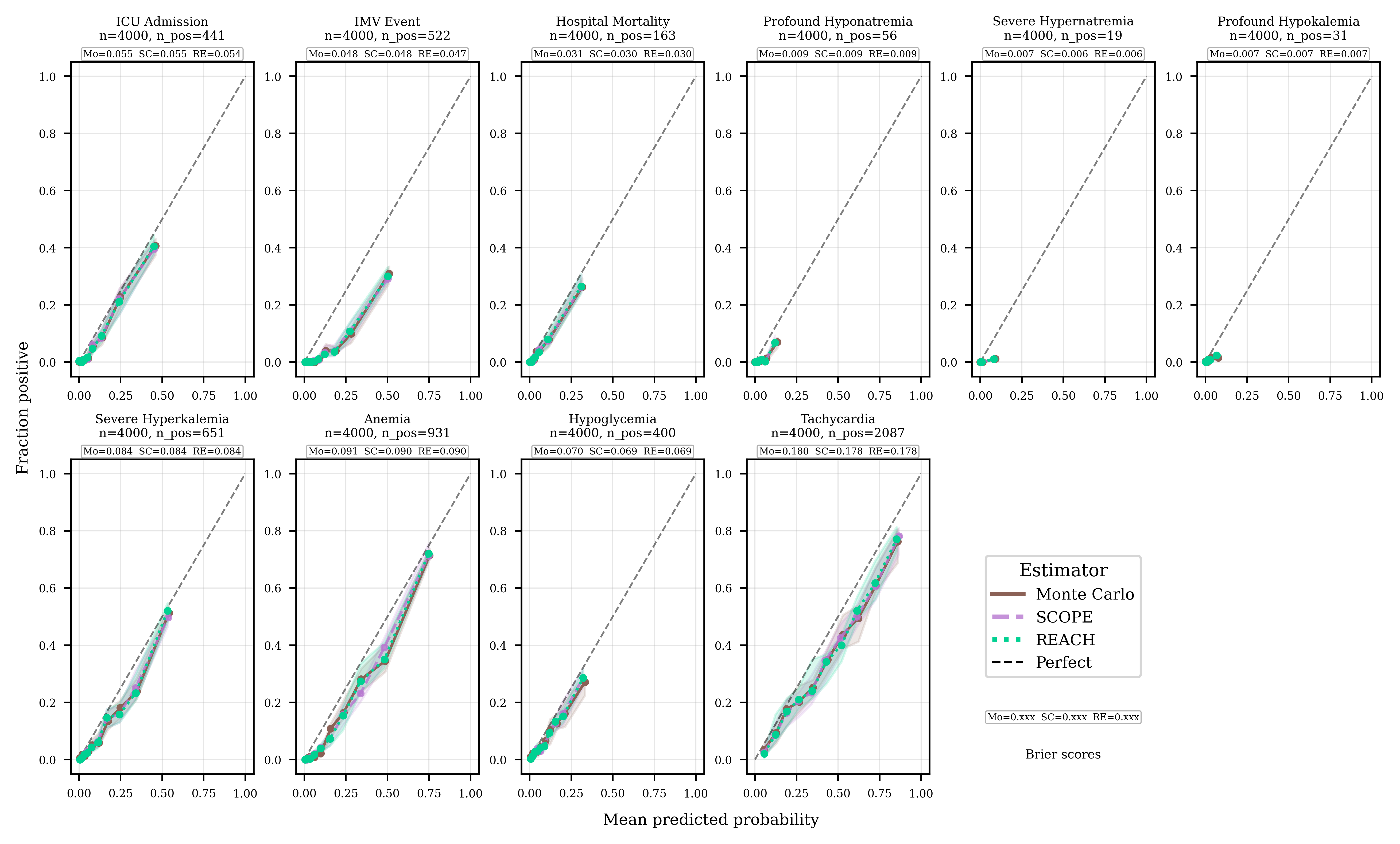}
    \caption{Brier scores and calibration curves of UCMC prediction tasks}
    \label{fig:placeholder}
\end{figure}
\clearpage

\section{Dispersion of Estimates Computation}
\label{app:dispersion}
Consider some outcome with general prevalence of $1/10000$. Next, we want to consider the probability that a patient with $10\times$ the general population risk is assigned a risk score by the Monte Carlo estimator that exceeds the risk score of a patient with risk equal to the prevalence.

The PMF for the Binomial distribution is
\[
f(k; n, p) = \binom{n}{k} p^k (1-p)^{n-k}.
\]

In order to compute the probability that the higher risk patient is ranked higher, we sum over all possible outcomes for the average risk patient and multiply each by the probability that the higher risk patient generates more samples that indicate the outcome of interest:
\[
\sum_{k=0}^{100} f\!\left(k; 100, \tfrac{1}{10000}\right) \cdot \sum_{j=k+1}^{100} f\!\left(j; 100, \tfrac{1}{1000}\right) = 0.0943.
\]
Thus there is only a $9.43\%$ chance that a patient with risk $10\times$ baseline is given a higher risk score than a patient with average risk.

\clearpage
\section{REACH is a Rao--Blackwellization}
\label{app:re_rao}
Let $(\hat X_{1:T},C_{1:T})$ be a pair of random variables corresponding to a backbone, so that $\hat X_{1:T}\sim \hat p$ is drawn from the outcome-avoiding distribution, $T=T_E(\hat X)$, and 
\[
C_t\sim \operatorname{Bernoulli}( h( \hat X_{1:t-1}))
\]
where $h(\hat x_{1:t-1}) = P(X_t = O | X_{1:t-1} = \hat x_{1:t-1})$ for all $1\leq t \leq T$. Note that each $C_t$ is independent after conditioning on $\hat X_{1:T}$. From the argument that $P(A) = P(B)$, it follows that
\[
P(T_O < T_E) = \EE[\1_{\{\exists t:\, C_t = 1\}}].
\]
Clearly, $\hat X_{1:T}$ is a sufficient statistic for each $C_t$ and thus all $C_{1:T}$, so that Rao--Blackwell's theorem implies that the following estimator has reduced variance:
\begin{align*}
\mathcal R_1 
&= \EE[\1_{\{\exists t:\, C_t = 1\}} | \hat X_{1:T}] \\
&= 1 - \EE[\1_{\{C_t \ne 1\, \forall t\}} | \hat X_{1:T}] \\
&= 1 - \textstyle \prod_{t=1}^T \EE[\1_{\{C_t \ne 1\}} | \hat X_{1:t-1}] \\
&= 1 - \textstyle \prod_{t=1}^T (1- h(\hat X_{1:t-1}))
\end{align*}
where conditional independence and the fact that 
\[
\EE[\1_{\{C_t \ne 1\}} | \hat X_{1:T}] = \EE[\1_{\{C_t \ne 1\}} | \hat X_{1:t-1}]
\]
justifies the third equality. REACH is the sampled version of $\mathcal R_1$.

\newpage
\section{REACH is Rao-Blackwellization of Arbitrary Rare-Event IS}
\label{app:reach_wins}
Consider some next token generation model that generates a probability distribution for the next token conditioned on the sequence up to this point. Given some arbitrary stopping condition, we want to estimate
$P(T_O < T_E)$.

If this outcome of interest token appearing before the end of the timeline is relatively rare event, one straightforward way to decrease the variance of a Monte Carlo sampling estimate is to perform importance sampling in order to increase the probability that the next token is the outcome of interest token during sampling.

Let $f$ denote the ordinary distribution of timelines. Assume that we have some arbitrary proposal distribution $g$ with the following restriction:

$$ P_g(X_{n}\  | X_{1:n-1} \cap O\not \in X_{1:n}) = P_f(X_{n}\  | X_{1:n-1} \cap O\not \in X_{1:n})$$

In other words, this class of importance sampling techniques only scales the probability that the outcome of interest token is generated. This is a reasonable restriction since there is no trivial choice of distribution that changes the probability of the non-outcome of interest tokens in a medical setting. 

Given this restriction, consider the alternative probability space $(\hat X_{1:n}, C_{1:n})$. $\hat X_{1:t}$ is drawn from the outcome avoiding distribution and $C_t \sim\text{Bernoulli}(P_g(X_t = O | \hat X_{1:t-1}))$. Note that the estimator now becomes: $\frac{1}{m}\mathbb{E}[\frac{f((\hat X_{1:T}, C_{1:T}))}{g((\hat X_{1:T}, C_{1:T}))} \mathbbm{1}_{C_i = 1}]$

By our above restriction, $\frac{f((\hat X_{1:T}, C_{1:T}))}{g((\hat X_{1:T}, C_{1:T}))} = \frac{f(C_{1:T})}{g(C_{1:T})}$.

Let $\Omega_1$ be the subset of the original sample space where the outcome of interest is sampled at some point. Let $\Omega_2$ be the subset of the seconds space where $C_i = 1$ for some $i$. Let $h:\Omega_1\rightarrow\mathcal{P}(\Omega_2)$ be the following mapping between $\Omega_1$ and the subsets of $\Omega_2$. 

(Note that for this argument I'm assuming sequences terminate at $T_O$) For each sequence $Y_{1:T_0}$ in the original sample space, map it to the set of all $(\hat X_{1:T}, C_{1:T})$ such that $Y_{1:T_0-1} = \hat X_{1:T_0-1}$ and $C_{1:T_O-1} = 0$ but $C_{T_0} = 1$. Note that the probability of $X_{1:T_0}$ and the probability of the event $h(X_{1:T_0})$ are equal:
\begin{align*}
    P_g(h(X_{1:T_0})) &=\frac{P_g(Y_1 =X_1)}{1 - P_g(Y_1 = O)}\\
    &\cdot\bigg(\prod_{t = 2}^{T_O-1} \frac{P_g(Y_t = X_t | Y_{1:t-1} = X_{1:t-1})}{1 - P_g(Y_t = O | Y_{1:t-1} = X_{1:t-1})}\bigg) P_g(Y_{T_0} = O | Y_{1:T_0 - 1} = X_{1:T_0-1})\\
    &\cdot(1-P_g(Y_1 = O) \prod_{t=2}^{T_O-1}(1 - P_g(Y_t = O | Y_{1:t-1} = X_{1:t-1}))\\
    &=P_g(Y_1 =X_1)\bigg(\prod_{t = 2}^{T_O-1} P_g(Y_t = X_t | Y_{1:t-1} = X_{1:t-1})\bigg) P_g(Y_{T_0} = O | Y_{1:T_0 - 1} = X_{1:T_0-1})\\
    & = P_g(X_{1:T_O})
\end{align*}
Next, note that by our restriction:
\begin{align*}
    \frac{f(h(X_{1:T_O}))}{g(h(X_{1:T_O}))} &=  \frac{P_f(X_1 \neq O)(\prod_{t=2}^{T_O-1}P_f(X_t \neq O | X_{1:t-1}))(P_f(X_{T_O} = O | X_{1:t-1}))}{P_g(X_1 \neq O)(\prod_{t=2}^{T_O-1}P_g(X_t \neq O | X_{1:t-1}))(P_g(X_{T_O} = O | X_{1:t-1}))}\\
    & = \frac{f(X_{1:T_O})}{g(X_{1:T_O})}
\end{align*}

Thus, we have shown that the this mapping maintains both probabilities and sample weights. Thus, we can conclude that the variance of the two estimators are equal.

Next, consider the Rao-Blackwellization using the outcome-free backbone as the sufficient statistic. By our earlier work, it is simple to show that this yields:

$$\hat p = \sum_{t=1}^{T}\frac{P_f(C_{1:t-1} = 0)P_f(C_t = 1)}{\cancel{P_g(C_{1:t-1} = 0)P_g(C_{t} =1)}}\cancel{P_g(C_{1:t-1} =0)\cdot P_g(C_{t} = 1)}$$

Which is exactly equal to the non-importance sampling REACH estimator. Since this was a Rao-Blackwellizaiton of an estimator with equal variance to the importance sampling estimator in the original space, this implies that REACH is a lower variance estimator for all possible choices of $g$ that do not alter the distribution of timelines conditioned on the non-appearance of the outcome of interest. 

\clearpage

\section{Tokenization}
\label{app:tokenize}
Tokenization proceeded as follows. Both MIMIC and UCMC data were converted to the Common Longitudinal ICU Format \citep[CLIF:][]{Roj25} version 2.1 standard that harmonizes critical care data across institutions by mapping site-specific laboratory and vital codes to a minimum set of common ICU data elements. Each hospitalization was assigned a sequence of tokens, starting with a \texttt{BOS} beginning-of-sequence token. Tokens for age, sex, race, ethnicity, and admission type were inserted in this order at the time of admission. Interdepartmental transfers within the hospital received one token each for ingress and egress, at their respective times. Medications with dosages, labs with quantitative results, and vitals with measured values each received category-value tokenization. For each category (standardized kind medication, type of lab, type of vital), all numerical values reported within the training set corresponding to that category were used to determine decile cutoffs. All values for that category (in training, tuning, and held-out data) were then binned into \texttt{Q0}, \texttt{Q1}, \dots, \texttt{Q9} according to those deciles~\cite[cf.][]{BBJ-preprint}. For each category-value pair, a new ``fused'' token~\cite{Lee26} was created (for example, \texttt{LAB-RES//so2\_arterial\_Q0} corresponds to a lab result for arterial SO$_2$ in the lowest decile). Tokens for medications and labs were inserted at the time of administration or measurement. For labs, a token corresponding to the lab was inserted at the time the lab was ordered, and two tokens correspond to both the lab and its binned numerical value were inserted at the time the result was available. Changes in code status were tokenized similarly. Continuous renal replacement therapy with blood flow rate, patient assessments with numerical value where appropriate, and respiratory support with fio2, peep, and tidal volume values also received category-value tokenization. Changes in code status, the practice of proning a patient, and discretized measurements of sofa score (as provided by the clifpy package) were also tokenized. All sequences terminate with an \texttt{EOS} end-of-sequence token. See Table~\ref{tbl:vocab} for further details on the tokens.

\begin{table}[h!]
\caption{A breakdown of the 1385-token vocabulary by token type}
\label{tbl:vocab}
\centering
\makebox[\textwidth][c]{
\begin{tabular}{lllr}
\toprule
category & description & example token(s) & count \\
\midrule
ADMN & admission type & \texttt{ADMN//direct}, \texttt{ADMN//elective} & 3 \\
AGE & age & \texttt{AGE//age\_Q0}, \texttt{AGE//age\_Q5}, \texttt{AGE//age\_Q9} & 10 \\
ASMT & assessment & \texttt{ASMT//braden\_activity\_Q6}, \texttt{ASMT//rass\_Q4} & 75 \\
CODE & code status & \texttt{CODE//dnr}, \texttt{CODE//full} & 4 \\
CRRT & dialysis & \texttt{CRRT//cvvh\_Q0}, \texttt{CRRT//cvvhd\_Q2} & 23 \\
DSCG & discharge & \texttt{DSCG//acute\_care\_hospital}, \texttt{DSCG//home} & 12 \\
ETHN & ethnicity & \texttt{ETHN//hispanic}, \texttt{ETHN//unknown} & 3 \\
LAB-ORD & lab order & \texttt{LAB-ORD//albumin}, \texttt{LAB-ORD//alt}, \texttt{LAB-ORD//wbc} & 45 \\
LAB-RES & lab result & \texttt{LAB-RES//albumin\_Q0}, \texttt{LAB-RES//alt\_Q2} & 473 \\
MED-CTS & continuous medication & \texttt{MED-CTS//dopamine\_Q8} & 388 \\
MED-INT & intermittent medication & \texttt{MED-INT//adenosine\_Q7} & 155 \\
POSN & proning & \texttt{POSN//prone} & 1 \\
RACE & race & \texttt{RACE//asian}, \texttt{RACE//unknown}, \texttt{RACE//white} & 7 \\
RESP & respiratory support & \texttt{RESP//fio2\_set\_Q1} & 35 \\
SEX & sex & \texttt{SEX//female}, \texttt{SEX//male} & 2 \\
SOFA & sofa score & \texttt{SOFA//cns-0}, \texttt{SOFA//coag-0}, \texttt{SOFA//resp-4} & 29 \\
TIME & time-spacing & \texttt{TIME//12h-1d}, \texttt{TIME//1w-2w}, \texttt{TIME//6mt+} & 13 \\
VTL & vitals & \texttt{VTL//dbp\_Q0}, \texttt{VTL//dbp\_Q5}, \texttt{VTL//weight\_kg\_Q9} & 88 \\
XFR-IN & transfer in & \texttt{XFR-IN//ed}, \texttt{XFR-IN//psych}, \texttt{XFR-IN//ward} & 8 \\
XFR-OUT & transfer out & \texttt{XFR-OUT//ed}, \texttt{XFR-OUT//psych}, \texttt{XFR-OUT//ward} & 8 \\
special & miscellaneous & \texttt{BOS}, \texttt{EOS}, \texttt{UNK} & 3 \\
\bottomrule
\end{tabular}
}
\end{table}

\clearpage
\section{Spontaneity and Prevalence Fit}
\label{app:spont_prev}

\begin{figure}[h!]
    \centering
    \resizebox{0.8\textwidth}{!}{%
        \includegraphics{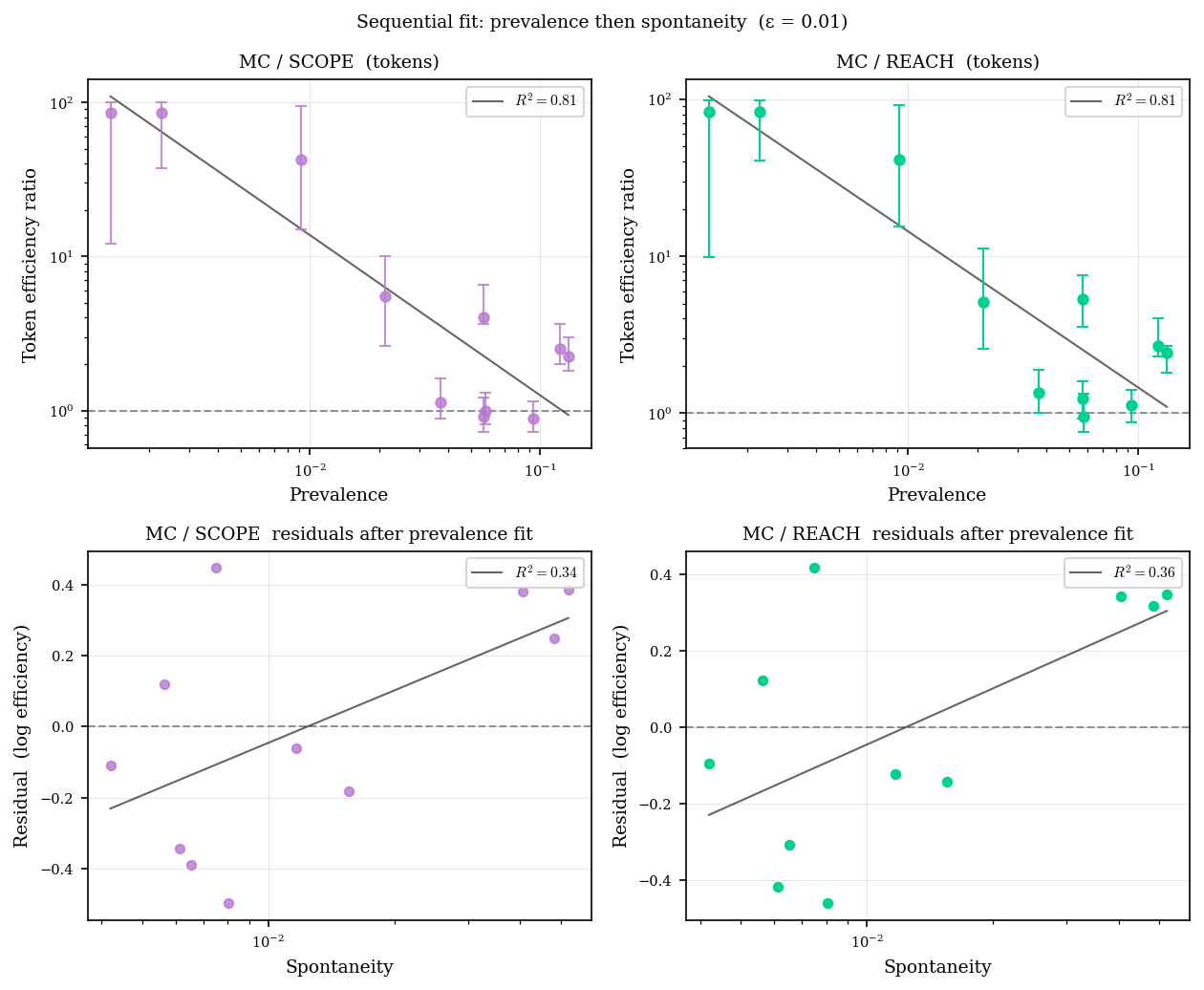}
    }
    \hspace{0.5em}
    \resizebox{0.8\textwidth}{!}{%
        \includegraphics{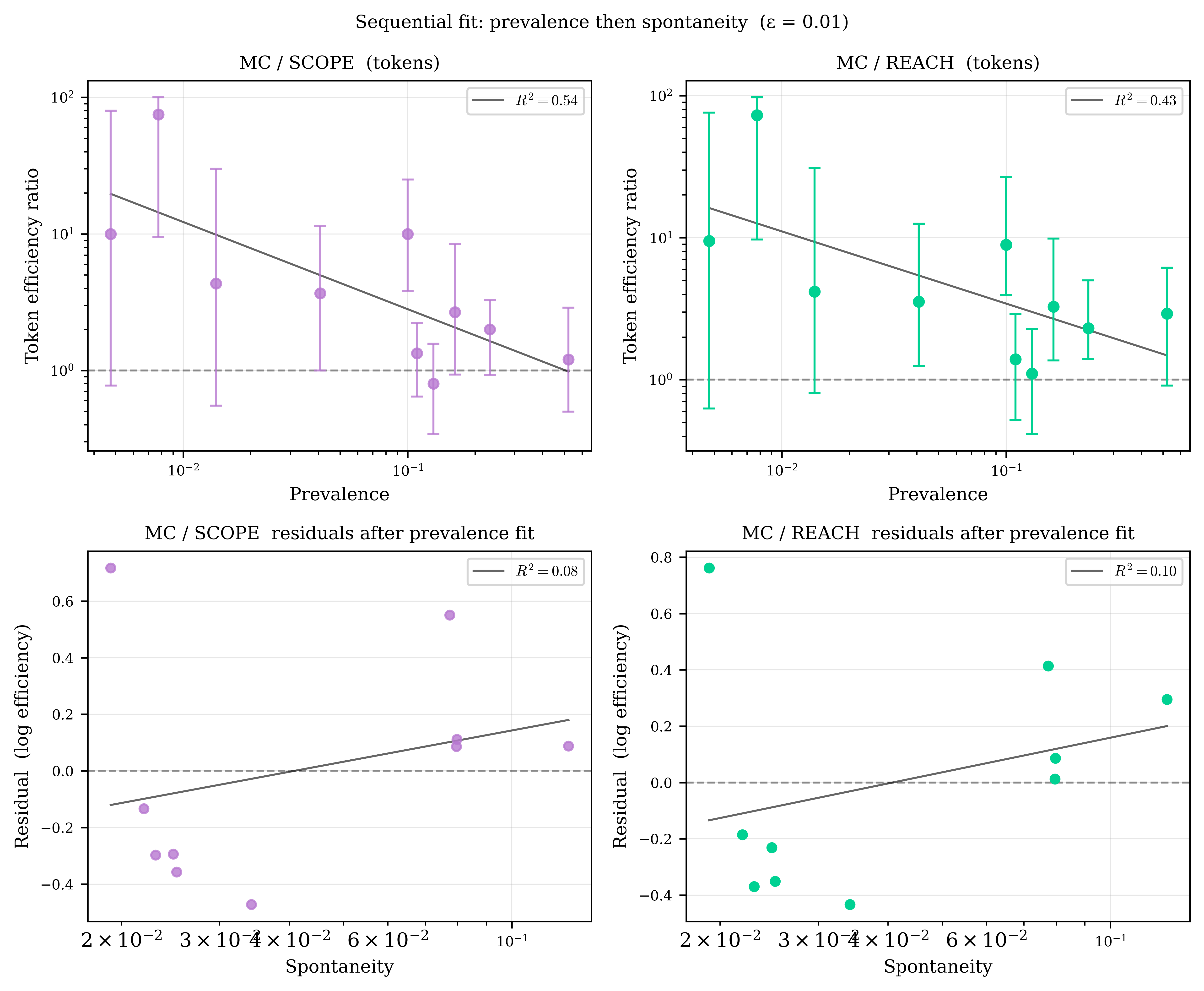}
    }
    \label{fig:myplots}
\end{figure}
\newpage
\section{Full Table 1}
\label{app:full_t_1}
\begin{table}[h!]
\centering
\caption{Token efficiency of SCOPE and REACH vs.\ Monte Carlo baseline at $\varepsilon=0.01$ on MIMIC-IV. Values are median token cost per patient [95\% bootstrap CI] across 100 replications. REACH cost is $M_1 + p \times M_2$. \textbf{Bold} indicates lowest token cost (where $M_1$ is mean standard generation length and $M_2$ is mean task specific REACH completion rate and $p$ is predicted prevalence). Ratios $>1$ indicate fewer tokens needed than MC.}
\label{tab:token_efficiency_mimic}
\small
\setlength{\tabcolsep}{3pt}
\makebox[\linewidth][c]{
\begin{tabular}{@{}l rrr cc@{}}
\toprule
\textbf{Outcome}
  & \boldmath$T_{\text{MC}}$ & \boldmath$T_{\text{SCOPE}}$ & \boldmath$T_{\text{REACH}}$
  & \textbf{MC/SCOPE} & \textbf{MC/REACH} \\
\midrule
ICU Admission
  & 30k [27k, 33k] & 30k [24k, 36k] & 31k [24k, 35k]
  & 1.00 [0.82, 1.31] & 0.95 [0.76, 1.34] \\
IMV Event
  & 33k [27k, 36k] & 36k [27k, 42k] & \textbf{26k [23k, 34k]}
  & 0.92 [0.73, 1.22] & 1.25 [0.93, 1.59] \\
Hospital Mortality
  & 33k [27k, 42k] & \textbf{5993 [3566, 12k]} & 6413 [3207, 13k]
  & 5.50 [2.62, 10.00] & 5.14 [2.57, 11.21] \\
Profound Hyponatremia
  & 51k [42k, 57k] & \textbf{1199 [599, 2996]} & 1240 [620, 3099]
  & 42.50 [15.00, 95.00] & 41.09 [15.47, 91.85] \\
Severe Hypernatremia
  & 51k [30k, 60k] & \textbf{599 [599, 4195]} & 611 [611, 5243]
  & 85.00 [12.04, 100.00] & 83.31 [9.80, 98.01] \\
Profound Hypokalemia
  & 51k [39k, 60k] & \textbf{599 [599, 1199]} & 613 [613, 1225]
  & 85.00 [37.50, 100.00] & 83.16 [40.30, 97.84] \\
Severe Hyperkalemia
  & 30k [27k, 33k] & 12k [8989, 15k] & \textbf{11k [7457, 11k]}
  & 2.50 [2.00, 3.67] & 2.68 [2.31, 4.02] \\
Anemia
  & 27k [24k, 30k] & 12k [8989, 15k] & \textbf{11k [11k, 15k]}
  & 2.25 [1.80, 3.00] & 2.42 [1.81, 2.68] \\
Hypoglycemia
  & 36k [33k, 42k] & 8989 [5993, 8989] & \textbf{6788 [4752, 10k]}
  & 4.00 [3.67, 6.50] & 5.30 [3.53, 7.57] \\
Tachycardia
  & 24k [21k, 27k] & 27k [21k, 33k] & \textbf{21k [17k, 26k]}
  & 0.89 [0.73, 1.14] & 1.12 [0.87, 1.40] \\
Positive Confusion
  & 27k [24k, 33k] & 24k [18k, 30k] & \textbf{20k [16k, 28k]}
  & 1.12 [0.89, 1.62] & 1.36 [1.01, 1.89] \\
\midrule
\multicolumn{1}{@{}l}{\textit{Aggregate:}}
  & & & & \textbf{SCOPE} & \textbf{REACH} \\
\quad Mean
  & & & & 20.97 [12.91, 26.98] & 20.71 [12.74, 26.68] \\
\quad Median
  & & & & 2.50 [2.25, 3.51] & 2.68 [2.42, 4.02] \\
\bottomrule
\end{tabular}
}
\end{table}

\begin{table}[h!]
\centering
\caption{Token efficiency of SCOPE and REACH vs.\ Monte Carlo baseline at $\varepsilon=0.01$ on the UCMC dataset. Values are median token cost per patient [95\% bootstrap CI] across 100 replications. REACH cost is $M_1 + p \times M_2$. \textbf{Bold} indicates lowest token cost. Ratios $>1$ indicate fewer tokens needed than MC.}
\label{tab:token_efficiency_local}
\small
\setlength{\tabcolsep}{3pt}
\makebox[\linewidth][c]{
\begin{tabular}{@{}l rrr cc@{}}
\toprule
\textbf{Outcome}
  & \boldmath$T_{\text{MC}}$ & \boldmath$T_{\text{SCOPE}}$ & \boldmath$T_{\text{REACH}}$
  & \textbf{MC/SCOPE} & \textbf{MC/REACH} \\
\midrule
ICU Admission
  & 73k [37k, 119k] & \textbf{55k [28k, 92k]} & 53k [32k, 84k]
  & 1.33 [0.64, 2.23] & 1.39 [0.52, 2.91] \\
IMV Event
  & 73k [37k, 115k] & 92k [50k, 170k] & \textbf{66k [38k, 133k]}
  & 0.80 [0.34, 1.57] & 1.11 [0.41, 2.28] \\
Hospital Mortality
  & 101k [55k, 156k] & \textbf{28k [9181, 83k]} & 29k [9513, 72k]
  & 3.67 [1.00, 11.43] & 3.54 [1.24, 12.55] \\
Profound Hyponatremia
  & 119k [64k, 174k] & \textbf{28k [3673, 174k]} & 29k [3822, 162k]
  & 4.33 [0.55, 30.00] & 4.16 [0.80, 30.87] \\
Severe Hypernatremia
  & 92k [15k, 174k] & \textbf{9181 [1836, 125k]} & 9670 [1934, 125k]
  & 10.00 [0.78, 80.00] & 9.49 [0.63, 75.96] \\
Profound Hypokalemia
  & 138k [28k, 184k] & \textbf{1836 [1836, 9181]} & 1889 [1889, 9447]
  & 75.00 [9.47, 100.00] & 72.89 [9.72, 97.18] \\
Severe Hyperkalemia
  & 73k [46k, 119k] & 28k [13k, 65k] & \textbf{22k [11k, 45k]}
  & 2.67 [0.93, 8.46] & 3.27 [1.36, 9.85] \\
Anemia
  & 55k [37k, 73k] & 28k [18k, 55k] & \textbf{24k [12k, 36k]}
  & 2.00 [0.93, 3.26] & 2.30 [1.40, 5.01] \\
Hypoglycemia
  & 92k [50k, 143k] & \textbf{9181 [3673, 23k]} & 10k [4122, 18k]
  & 10.00 [3.83, 25.00] & 8.91 [3.93, 26.73] \\
Tachycardia
  & 55k [28k, 101k] & 46k [28k, 83k] & \textbf{19k [13k, 40k]}
  & 1.20 [0.50, 2.88] & 2.92 [0.91, 6.14] \\
\midrule
\multicolumn{1}{@{}l}{\textit{Aggregate:}}
  & & & & \textbf{SCOPE} & \textbf{REACH} \\
\quad Mean
  & & & & 11.10 [4.34, 19.05] & 11.00 [4.81, 19.03] \\
\quad Median
  & & & & 3.17 [1.45, 5.50] & 3.41 [1.89, 6.03] \\
\bottomrule
\end{tabular}
}
\end{table}


\end{document}